\documentclass[12pt]{report}
\usepackage{url}
\usepackage{graphicx, color}
\usepackage[utf8]{inputenc}
\usepackage[english]{babel}
\usepackage[a4paper,margin=2cm]{geometry}
\usepackage[nonumberlist]{glossaries}
\usepackage{svg}
\usepackage[most]{tcolorbox}
\usepackage{algorithm}
\usepackage{subcaption}
\usepackage{algorithm}
\usepackage{algpseudocode}  
\usepackage{appendix}
\usepackage{multirow}
\usepackage{float}
\makeglossaries

\tcbset{
  notetoreader/.style={
    colback=green!3!white,       
    colframe=green!30!gray,      
    title=Note to the reader,    
    fonttitle=\bfseries,         
    coltitle=black,              
    boxrule=0.6pt,               
    arc=3mm,                     
    left=6pt,                    
    right=6pt,                   
    top=6pt,                     
    bottom=6pt                   
  }
}
\newglossaryentry{metalm}
{
    name=metal mesh,
    description={Material like a net with spaces in it, made from metal wire},
    plural={metal meshes},
    sort={2}
}
\newglossaryentry{mathm}
{
    name=mathematical mesh,
    description={A set of points connected by edges, representing the surface of some object},
    plural={mathematical meshes},
    sort={1}
}
\newglossaryentry{numsurr}
{
    name=surrogate model,
    description={A computer program that can be used instead of something else (such as another computer program)},
    sort={3}
}
\newglossaryentry{phantom}
{
    name=phantom,
    description={A physical object mimicking a part of a body, that can be used instead of the real part (for training or experiments)},
    sort={4}
}
\newglossaryentry{physout}
{
    name=physical outcome,
    description={What literally happens with all the objects involved in a medical procedure (such as how the blood clot moves when doing mechanical thrombectomy)},
    sort={5}
}
\newglossaryentry{clinout}
{
    name=clinical outcome,
    description={What happens to the medical condition of a patient in the hours, months or years after a medical procedure (such as death, or full recovery)},
    sort={6}    
}
\newglossaryentry{ANN}
{
    name=artificial neural network,
    description={a machine learning model that uses parameters learned on data, linear algebra and non-linear functions to predict outputs given inputs},
    sort={7}
}
\newglossaryentry{BNN}
{
    name=biological neural network,
    description={A set of connected neuron cells ex. the brain},
    sort={8}
}

\begin{document}
\renewcommand*{\glstextformat}[1]{\textit{#1}}
\renewcommand{\glossarysection}[2][]{%
  \section*{#2}%
}

\begin{titlepage}

\newcommand{\HRule}{\rule{\linewidth}{0.5mm}} 

\center 




\textsc{\Large MSc Artificial Intelligence}\\[0.2cm]

\textsc{\Large Master Thesis}\\[0.5cm]




\HRule \\[0.4cm]

{ \huge \bfseries An Exploratory Study into using Machine-Learning for Fast Step-by-step Emulation of Numerical Mechanical Thrombectomy Simulations for Ischemic Stroke}\\[0.4cm] 

\HRule \\[0.5cm]




by\\[0.2cm]

\textsc{\Large Thijs Stessen}\\[0.2cm] 

15159825\\[1cm]




{\Large \today}\\[1cm] 




\begin{minipage}[t]{0.4\textwidth}

\begin{flushleft} \large

\emph{Supervisor:} \\

MSc. Thijs Kuipers

\end{flushleft}

\end{minipage}

~

\begin{minipage}[t]{0.4\textwidth}

\begin{flushright} \large

\emph{Examiner:} \\

Dr. Simone Saitta \\

\vspace{0.5cm}

\end{flushright}

\end{minipage}\\[2cm]




\includegraphics[width=10cm]{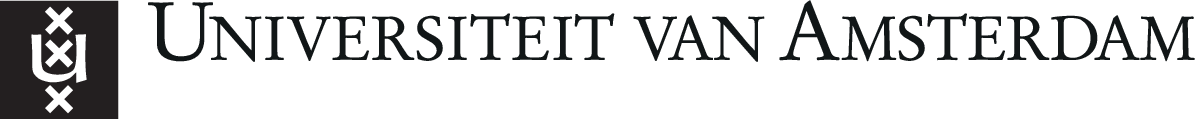}


\vfill 

\end{titlepage}
\begin{tcolorbox}[notetoreader]
This thesis contains terms from Physics, Medicine and AI. In some cases these terms overlap and mean different things depending on which field they are from, in which case we have split them up using an appropriate adjective (such as splitting 'mesh' into 'metal mesh' and 'mathematical mesh'). In other cases very similar concepts are used under different names, making it confusing what the differences are. When either of such cases occur we have written the words in \textit{italics} and added an entry to the glossary.
\end{tcolorbox}

\printglossaries
\pagenumbering{roman}

\tableofcontents

\begin{abstract}
The treatment of ischemic stroke using mechanical thrombectomy involves difficult decisions under intense time constraints. Numerical physics simulations can in theory inform operators to make better decisions regarding treatment approaches and device selection, but are too slow to do so in practice. In this thesis, we investigate if current machine learning based surrogates can accurately emulate these simulations in a step-by-step manner while making them significantly faster. To do this we train three surrogate models on two simulations that involve a simplified aspiration procedure, with varying levels of geometric complexity. Our results show that two of our models  accurately predict singular simulation steps and provide substantial speedups, especially when combined with specific data augmentations. However, the models showed a lack of stability when emulating simulations with complex geometries over longer time periods. Overall, this work provides a foundation for future studies to develop
stable methods that scale to realistic numerical physics simulations of mechanical thrombectomy.
\end{abstract}

\pagenumbering{arabic}

\chapter{Introduction}

Acute Ischemic stroke (AIS), caused by a blood clot blocking an artery in the brain, results in the death of more than 3.2 million people per year globally as of 2019 and is projected to cause 4.9 million deaths by 2030 \cite{fan2023global}. Treatment of AIS is often not successful; in the Netherlands, around 20 percent of patients undergoing endovascular treatment die within 90 days, 48 percent never fully recover neurologically and only 32 percent achieve a good \gls{clinout}, i.e. they recover most of their neurological capabilities \cite{jansen2018endovascular}. In the United States, this 90 day mortality is around 14 percent \cite{anand2021trends} (although this is only based on privately insured individuals, which introduces a selection bias). Treatment of AIS broadly follows the following steps: imaging the brain using methods such as computed tomography (CT); administering a drug into the bloodstream that is meant to dissolve blood clots (intravenous thrombolysis with alteplase); if the patients health allows and if the blood clot is blocking a large vessel such as the middle cerebral artery or the intracranial carotid artery, physical removal of the blood clot via mechanical thrombectomy \cite{mosconi2022treatments}. 

Mechanical thrombectomy involves bringing a device through the patients vascular system to the location of the blood clot and then deploying it. This device is either an aspiration device which sucks in the blood clot, a stent retriever which encapsulates the blood clot using a physical net-like \gls{metalm} or a combination of both \cite{munich2019overview}. 
 Many different designs exist but in practice, the decision which mechanical thrombectomy device to use (from the selection available at that hospital) is left to the discretion of the operator, as there is currently no guideline on which approved device to use.
 
When mechanical thrombectomy is performed, the amount of attempts taken to remove the blood clot must be minimized to achieve the best \gls{clinout}. In fact, when only one attempt is needed a so called first pass effect is observed and the chance of a good \gls{clinout} is much higher \cite{zaidat2018first}. Moreover, each attempt comes at the risk of the defragmentation of the blood clot, possibly resulting in the fragments blocking other smaller blood vessels in the brain \cite{kaesmacher2017risk} which cannot be removed by mechanical thrombectomy. It is thus crucial to select a mechanical thrombectomy approach that offers a high chance of success while minimizing clot fragmentation.

How successful an attempt is and if fragmentation occurs, which we will call the \gls{physout}, is depended on the device used and the exact geometry of the blood vessel \cite{yeo2019does}.
However these two aspects also impact the \gls{physout} interdependently: some devices perform better in some geometries. For example, a stent retriever that is too small can create shearing of the blood clot, possibly creating small fragments or not fully removing the clot \cite{yeo2019does}. This might seem like a trivial example: simply choose the stent with the diameter that is closest to that of the blood clot. However, the choice of the correct approach is more nuanced in the case that the blood clot is of an intermediate size. Is a slightly larger or a slightly smaller stent better in that case? Additionally, when mechanical thrombectomy attempts fail multiple times, the operators commonly (around 80 percent of the time in one study \cite{Lapergue2017ASTER}) switch to a different approach (rescue therapy). This rescue therapy is mostly done using mechanical thrombectomy with a different device, showcasing that operators believe that another device will work better for the same patient (i.e. they believe a different device works better even though the geometry of the patients vascular system is exactly the same). If they did not believe so, they would not try mechanical thrombectomy again, due to the associated risks. Thus we will work under the idea that choosing the right device depending on the patient is likely to improve the success rate of each attempt.

 One of the most promising approaches that has been used to investigate the use of mechanical thrombectomy devices on different blood vessel geometries is the use of digital twins based on numerical physics simulations \cite{daei2023computational,luraghi2021first,liu2022simulation}. This approach involves creating a 3D model of a blood vessel, using the scan of a patient's vascular system (including the blood clot) and then using a numerical solver that incorporates known physical laws to simulate what would happen if a certain device was used to extract the blood clot. This simulation can then be used to analyze the likely \gls{physout} of the procedure. 
 Importantly, this simulation is specific to the patient. 
 However, there is a big disadvantage of using numerical physics simulations: they are very slow. Simulations with the accuracy required for the complex dynamics of blood clots and mechanical thrombectomy devices can take hours or even days on a computer cluster. In the clinical setting of acute ischemic stroke such time is not available since swift action (on the timescale of hours) is highly correlated with good \gls{clinout} \cite{groan2021time}. This is exacerbated by the fact that the time in which the simulation could be performed, the time between diagnosis using imaging and the start of the thrombectomy procedure, is even shorter, around 20-60 minutes \cite{musuka2015diagnosis}. In this paper, we attempt to tackle this problem of slow numerical simulations by replacing them with a \gls{numsurr} that uses machine learning to predict the next timestep instead, which has the potential to be much faster and robust to noise, which has been done previously on different tasks \cite{li2021neural,hemmasian2024multi}.

\begin{figure}
    \centering
    \includegraphics[width=\linewidth]{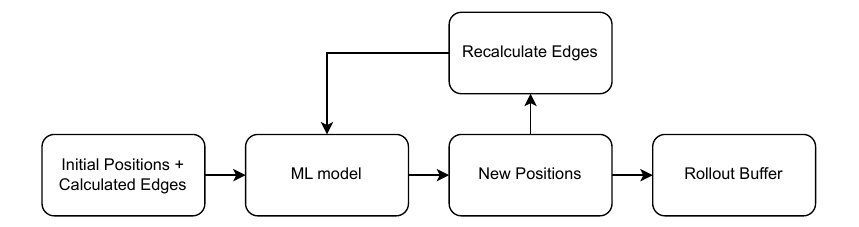}
    \caption{The step-by-step approach using a machine learning model as a \gls{numsurr}. The Rollout buffer can be made into a video or any other way to view a series of 3D points.}
    \label{fig:surrogate}
\end{figure}
The current machine learning based surrogates that have been used in the setting of ischemic stroke (or very related settings) are end-to-end methods. They take in some starting position, such as a 3D model of a blood vessel, and directly predict some feature of the result. These models can still be seen as \glspl{numsurr}, since they replace the numerical physics simulation. This has been done to directly predict the \gls{clinout}, with varying results \cite{daei2023computational}\cite{nishi2019predicting}. Other approaches rather try to predict part of the \gls{physout}, such as the stresses on the blood clot or on the blood vessel \cite{liang2017machine}. While useful for assessing fragmentation risk, they give little insight into exactly what caused this risk. They show the user a likely outcome, but do not provide insight into how this outcome came to be. Additionally, they give little insight into how the procedure could be modified to achieve a better outcome. To overcome this, we intend to use a \gls{numsurr} of the numerical simulation in a step-by-step fashion, see figure \ref{fig:surrogate}. This can be done in an iterative approach: the machine learning model is given the current state of the system and predicts the next one in a loop until some end time is reached. 
If our \glspl{numsurr} accurately emulates the full numerical simulation then this allows us to view the simulation in the same manner as if it was fully made numerically, while benefiting from the speedup and the resistance to noise that machine learning methods bring.

\begin{figure}
    \centering
    \includegraphics[width=\linewidth]{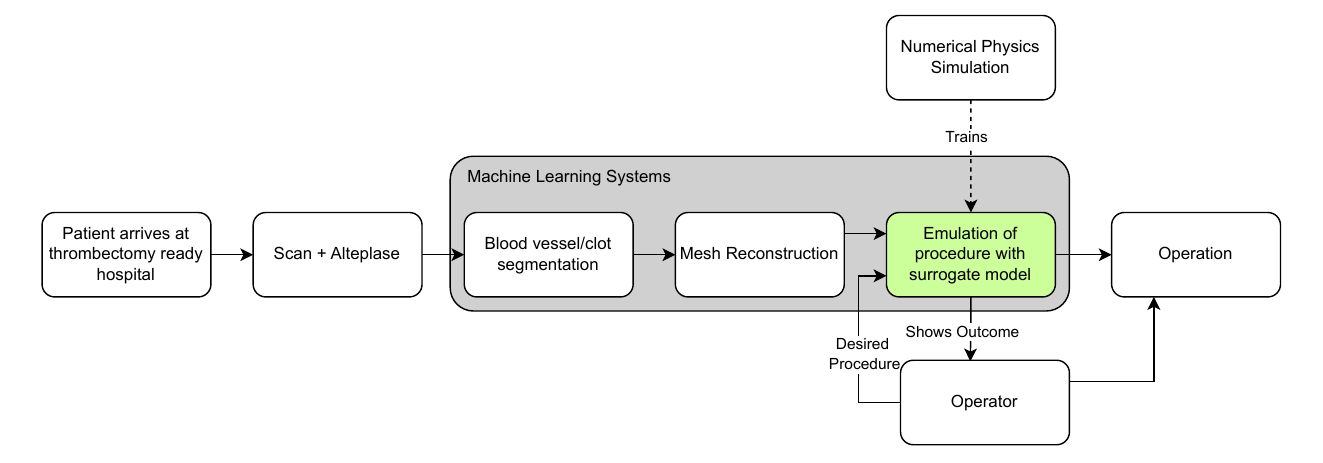}
    \caption{Overview of a possible pipeline of the use of digital twins in ischemic stroke treatment. Green represents the contributions of the current paper. Segmentation and reconstruction can currently be done by hand, but machine learning based models are being developed to automate this aspect.}
    \label{fig:overview}
\end{figure}

Currently, several machine learning methods exist that are suited for a step-by-step approach, however they have to be adjusted to the specifics of mechanical thrombectomy simulations, which will be the main focus of this thesis. For an overview of how this research fits into the future of ischemic stroke treatment, see figure \ref{fig:overview}.
The most promising machine learning methods, those that have been designed for a step-by-step approach and have been tested on datasets of physical systems, are either graph based or based on the attention mechanism. Graph-based methods include MeshGraphNet \cite{pfaff2020learning}, a graph \gls{ANN} with residual connections. Attention based methods include the Transolver \cite{wu2402transolver} and Erwin \cite{zhdanov2025erwin}. The main difference between these is how they handle inputs with large amounts of nodes and how their embeddings are calculated. We will investigate how these models compare in this setting, in terms of inference speed and in single-step and multi-step accuracy. Additionally, we will investigate methods for improving performance, such as generic data augmentations, but also data generation specific to 3D point cloud data. Namely, we will describe an algorithm for sampling new datapoints, by sampling on the triangles between three datapoints. Lastly, we will do two small ablation experiments, one to study the impact of data augmentation on generalizability, the other to study the impact of data-point size on speed and performance. This then gives us the following research questions:

\begin{itemize}
    \item How well are MeshGraphNet, Transolver and Erwin suited as step-by-step surrogate models for numerical mechanical thrombectomy simulations? \begin{itemize}
        \item How accurately do these models predict the next step of the numerical simulations?
        \item How stable are the predictions of these models across longer timescales?
        \item How fast is inference for these models compared to numerical solvers?
    \end{itemize}
    \item What is the impact of data augmentations on single-step performance?
    \begin{itemize}
        \item How does the addition of different amounts of generated datapoints to the training set impact performance?
        \item How does rotating the input data affect performance?
        \item How does increasing data-point size affect performance?
    \end{itemize}
    \item How does rotating the input data affect generalizability to unseen geometries?
\end{itemize}

To investigate these claims we use 2 novel datasets provided by inSteps B.V. which both involve a blood clot moving through a blood vessel. These datasets are build using a traditional solver (Abaqus) for numerical physics and involve predicting the next timestep given the current one. The first dataset, called BendVessel, resembles a simplified aspiration procedure in which a force is applied to remove the blood clot from a simplified blood vessel. The second dataset, called ClotEntry, is similar to the first, but has a much higher resolution and uses realistic blood vessels.
On these datasets we then train Meshgrapnet, Transolver and Erwin, adjusting the models and data as required. We investigate their performance on a single step and on a rollout and record their inference speed. Additionally, we investigate the use of various data augmentations and their effect on performance. Finally, we perform an experiment on generalizability to unseen geometries. 

\glsadd{BNN}


\chapter{Background}
This chapter introduces the concepts from Medicine, Numerical Physics and AI required to understand the context of this thesis and its methods. First, we will cover the basics of Ischemic Stroke, focused on the aspects that are most relevant for modeling it in numerical simulations. Second, we will introduce numerical physics simulations, how they work and how they relate to Medicine and AI. Thirdly, we will describe the existing Machine Learning methods that are relevant to this thesis. We will cover both end-to-end methods previously used in the mechanical thrombectomy setting as well as methods suitable for a step-by-step approach. Lastly, we describe an algorithm for sampling new data from step-by-step pointcloud datapoints.
\section{Ischemic stroke}
Ischemic stroke results in the loss of blood flow to parts of the brain, resulting in damage to the affected tissue. In the following chapter we will describe in more detail the aspects of this process that are most relevant for our purposes. Namely, we will describe the blood vessels that commonly get blocked, the nature and composition of blood clots that block them and the types of devices used to remove blood clots.

\subsection{Blood Vessels in the Brain} \label{bloodvessel}

Blood Clots in Ischemic Stroke cases can be removed mechanically if they are located in large vessels, which will thus be the vessels we will focus on. These Large Vessel Occlusions (LVO) occur in around $33\%$ of ischemic stroke cases and mostly ($70\%+$ of LVO's) occur in either the middle cerebral artery (M1) and its branches or the intracranial carotid artery. \cite{beumer2016occurrence}. These vessel are connected to the Circle of Willis, a circular system of blood vessels that can provide alternative routes for blood flow. Its shape is very variable \cite{kapoor2008variations}, so much so that the amount of connections between blood vessels can differ from person to person. These variations can also influence the blood flow into the M1 and the intracranial carotid artery \cite{sturiale2024geometry} and might thus be relevant for the interaction between blood clots in LVO's and the blood surrounding them.
These blood vessels branch into many smaller vessels. Some branches, such as the M2 which branches off the M1, are large enough for mechanical thrombectomy, while many of these branches are too narrow for the current methods \cite{jadhav2021indications}\cite{onal2020feasibility}.
\begin{figure}[H]
    \centering
    \includegraphics[width=0.5\linewidth]{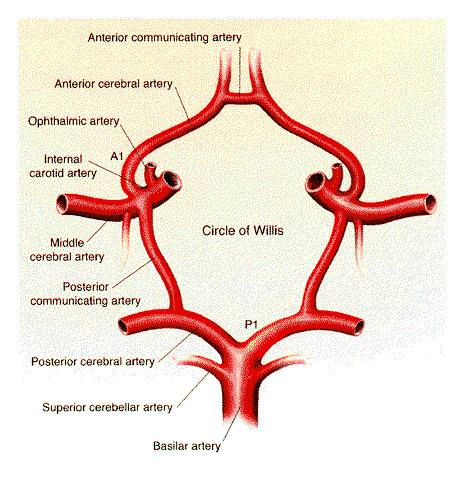}
    \caption{Schematic view of a typical example of the circle of Willis. The Middle Cerebral Artery (M1) and the Internal Carotid Artery (ICA), which is where most mechanical thrombectomy takes place, can be seen in the middle, on the left. Taken from \cite{schomer1994anatomy}}
    \label{fig:willis}
\end{figure}
For our purposes of modeling mechanical thrombectomy an important aspect of the blood vessels is their rigidity. During normal blood flow and during operations forces are applied to the blood vessel and change its shape. For example, when performing mechanical thrombectomy, the blood clot is pulled from its position while it is in contact with the blood vessel, creating friction. However, in all current 3D numerical physics simulations of mechanical thrombectomy it is assumed that these forces are insignificant compared to how rigid the blood vessels are \cite{daei2023computational}\cite{luraghi2021first}. This is based on the fact that the blood clots are much more elastic than the vessel walls (1-2 MPa vs. 10 kPa elastic modulus for the clot and vessel, respectively).

\subsection{Blood Clots}
Blood clots come in different sizes, shapes, and compositions which affect the their physical dynamics during treatment. We will discuss how these different factors impact the \gls{physout} of mechanical thrombectomy and what the limitations are of current simulations.

\begin{figure}[H]
    \centering
    \includegraphics[width=\linewidth]{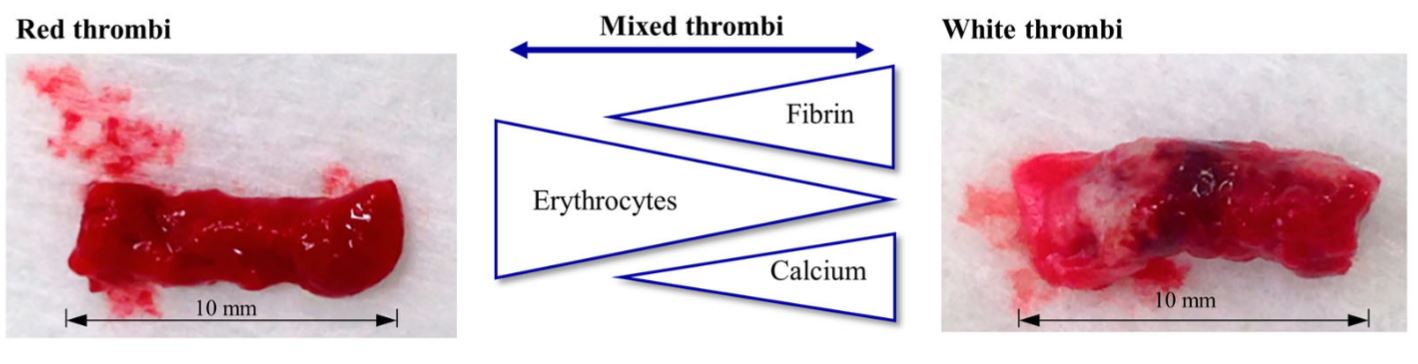}
    \caption{Different types of blood clots (thrombi). Calcium binds to the components of the blood clot and is more prevalent with increased platelets. The term Erythrocytes refers to red blood cells.  Taken directly from: \cite{wortmann2022development}.}
    \label{fig:bloodclots}
\end{figure}
Blood clots mainly consist of fibrin, platelets, red blood cells, white blood cells, Von Willibrand Factor (a large protein present in the blood) and extracellular DNA \cite{jolugbo2021thrombus}. All of these components occur in different ratios in different blood clots, causing them to be heterogeneous. Moreover, there is variation within singular blood clots, it is common that sections of a blood clot differ in their composition, for example a common distribution is a core high in red blood cell content and a shell with more fibrin, platelets and Von Willebrand Factor. 
\cite{di2019acute}.
The composition of a blood clot affects the likelihood of successful treatment using current mechanical thrombectomy procedures \cite{jolugbo2021thrombus}. In general, blood clots with more fibrin are stiffer and more difficult to remove, in contrast with those with more red blood cells, which are the opposite. In addition, increased fibrin content increases the friction the blood clot experiences with the blood vessel walls and therefore how it behaves during mechanical thrombectomy. Lastly, increased red blood cell content increases the risk of fragmentation of the blood clot, which can result in other blood vessels getting blocked.

The size of blood clot certainly has some relation to the \gls{physout}, but the exact nature of this relation is still an area of active research \cite{yeo2019does}. Larger clots are easier to remove, but do not lead to better \glspl{clinout}. This indicates that larger clots have a bigger chance of fracturing.

\subsection{Mechanical Thrombectomy Devices}

\begin{figure}[H]
    \centering
    \begin{subfigure}{0.8\textwidth}
        \centering
        \includegraphics[width=0.5\linewidth]{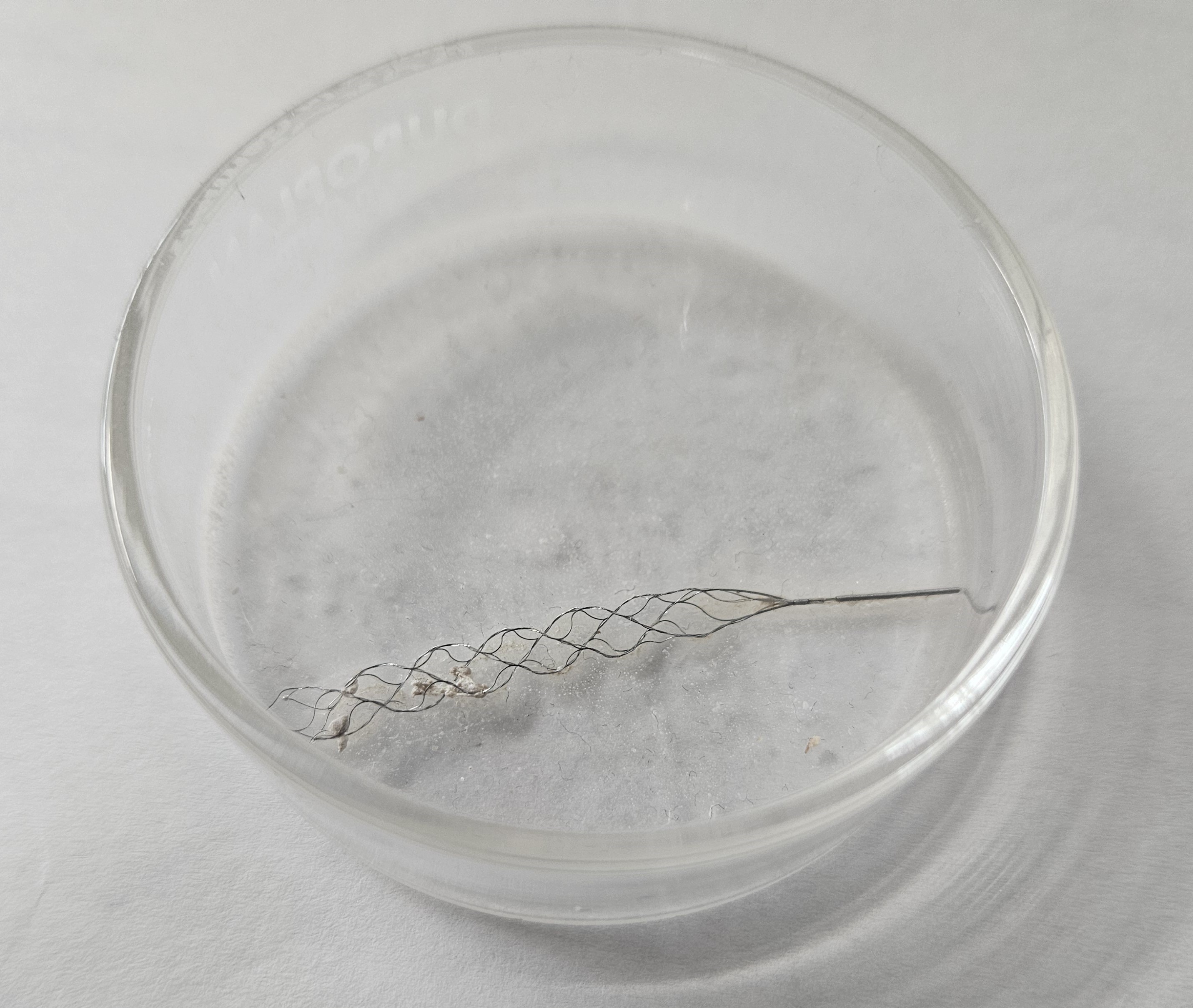}
        
    \end{subfigure}

    \vspace{0.5cm} 

    \begin{subfigure}{0.8\textwidth}
        \centering
        \includegraphics[width=\linewidth]{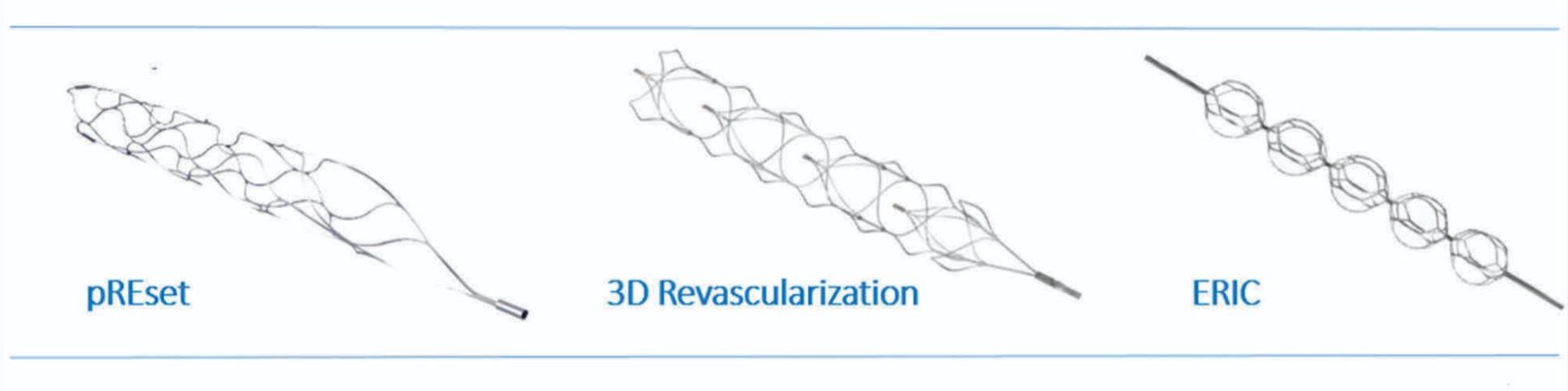}
        
    \end{subfigure}
    \begin{subfigure}{0.79\textwidth}
        \centering
        \includegraphics[width=\linewidth]{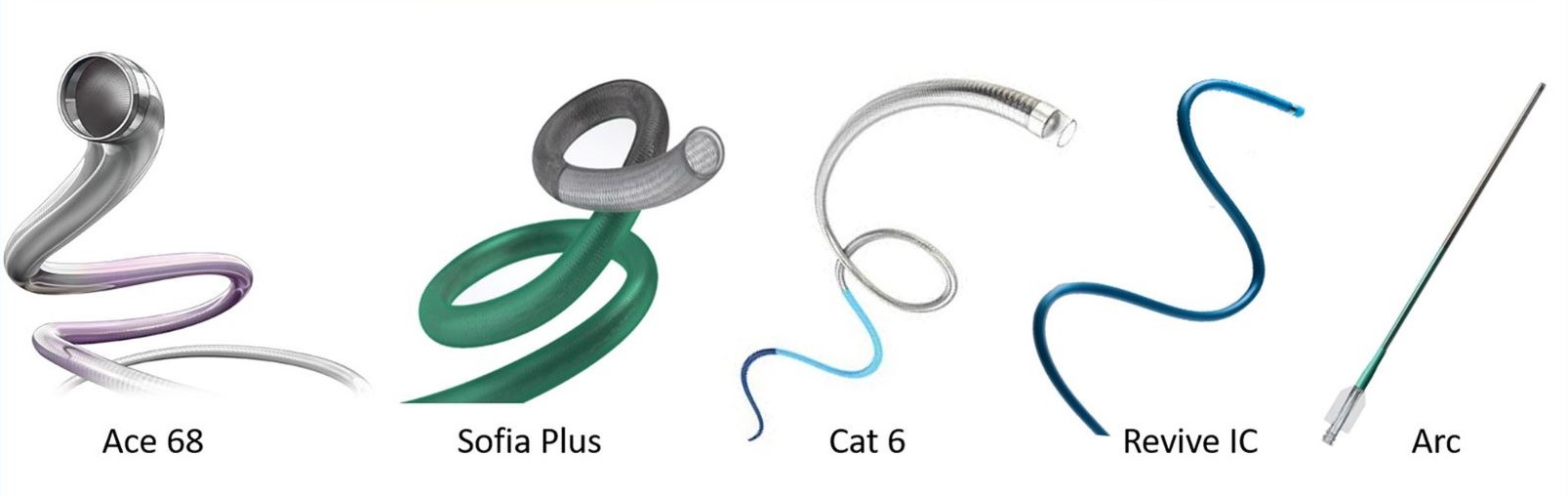}
        
    \end{subfigure}

    \caption{First picture: used stent retriever. Second picture: Example stent retriever and aspiration designs, adapted from \cite{gg2018acute}.}
\end{figure}
Mechanical thrombectomy devices come in two main types, aspiration devices and stent retrievers (or a combination of both). These devices are delivered to the location of the blood clot by following a guide wire that is inserted into the bloodstream. For stent retrievers, a microcatheter is used to facilitate this delivery \cite{blanc2020recent}. Many designs of these devices exist, varying in size and shape, for a comprehensive list see \cite{blanc2020recent}.

Aspiration devices function by applying a negative pressure to a hollow tube that is connected to the device. The operator does this by using a syringe manually or an electric pump. The blood clot is then sucked into device, either traveling through the tube and outside the body, or is stuck to tip and is removed along with the device. These devices vary in the design of their tips and in their size. Examples of aspiration devices include Penumbra Ace 68 \cite{penumbraACE} and Medtronic React 68 \cite{medtronicReactCatheter}.

Stent retrievers consist of \glspl{metalm} designed to grab hold of the blood clot. They are delivered to the blood clot in a collapsed state inside a sheath, where they puncture the blood clot. Once inside they exit the sheath and expand. When expansion is complete the stent retriever is pulled, along with the blood clot. Stent retrievers come in many different designs where not only their size, but the entire structure of their mesh varies. Examples of stent retrievers include Stryker Trevo XP \cite{strykerTrevoXP} and Terumo Neuro Eric \cite{terumoERIC}.

\section{Numerical Physics Simulations}
Computer simulations of physical processes allow the inspection of and the experimentation with physical systems in ways that would be infeasible or expensive otherwise. They do this by applying known physical laws to the state of a system to find the state that follows it. This then allows for inspection of details or failures by viewing the sequence of states and for experimenting with designs by changing the initial state or the simulation parameters. The use of physics simulations in these ways dates back to the 1950's \cite{rizzo1967integral} and they have since been used in applications across many subfields of engineering and physics like in applied mechanics \cite{belytschko1984explicit}, rocket engine design \cite{chen2011multiphysics} and nuclear physics \cite{langanke1991computational}, including in the intersection with the medical domain \cite{fumagalli2024role}.

In this section we will describe how numerical physics simulations work and what decisions are required when building one after which we will describe their advantages and disadvantages in the medical domain.
Finally, we will showcase the ways in which AI can be used to improve simulations and which challenges arise when doing so.

\subsection{Defining a Simulation}
In the simplest terms, a numerical physics simulation needs a start state at time $t=0$: $s_0$ and a transition function $F(s_t)$ with gives the state at time $t+\Delta t$, where $\Delta t$ is some small timestep. Then $F$ can be applied iteratively to $s_0$ to produce an entire sequence of states. To make it clear how the simulations described in the method section came to be we will now discuss the process of defining a state and a transition function and what decisions need to be made along the way.
\subsubsection{State}
The form that the state takes must be carefully considered as it has a large impact on the simulation as a whole. For some systems defining the state is straightforward, a ball flying through the air could be modeled as an $x,y,z$ position and velocity (although even here, polar coordinates could be considered as an alternative). For systems with many moving elements, such as fluid simulation or the simulation of galaxy formation an important decision is if the state should be discretized or not. A common approach in fluid simulation is to divide up the space into squares or boxes and to only model the interaction between these while ignoring any interactions on a smaller scale. This allows for fast computation, but comes at the cost of accuracy. A related approach is to treat surfaces as a mesh, with points sampled on the surface that are connected based on distance. This approach has the advantage of the fast computation, but for the right problem it suffers less from the loss in accuracy than the box approach. This is because it discretizes only the points of interaction, but not space itself, as the points can still have real valued positions. 

\subsubsection{Transition function}
Finding the transition function entails the following steps: finding the derivatives governing the system, deciding how to use those derivatives, filling out the other parameters in the derivatives and defining boundary conditions.

Firstly, when considering all the forces involved in a physical process we receive a set of equations that govern how the system behaves. These can take two forms, either an explicit form of a derivative of $s$ over time, or an implicit form of a system of equations that the current and future states must satisfy. 

Secondly, we then need to take these sets of equations and compute the next state. The best way to do this is very problem dependent. For the explicit case, the simplest method, known as the forward Euler method looks as follows:
\begin{equation}
s_{t+\Delta t} =F(s_t) = s_t + \Delta t \frac{d}{dt}s_t .
\end{equation}

 It must be noted that in many systems forward Euler is a very unstable approach and it will often diverge from the correct values. Several other methods exist, like the Leap Frog method and a family of methods known as Runge-Kutta methods, which also incorporate higher order derivatives. For the implicit case, the backward Euler method involves solving the following equation: \begin{equation}
    s_{t+\Delta t} = s_t + \Delta t \frac{d}{d(t+\Delta t)}s_{t+\Delta t}
\end{equation} 

Thirdly, we need to fill in some parameters present in the sets of equations. Some generic parameters might already be known, such as the thermal expansion coefficient of water. However, others will have to be measured to achieve accurate results. These could include the exact geometry of the starting state or exact physical conditions like air pressure or friction coefficients. Additionally, the size of $\Delta t$ is very important, as the larger it is the faster the simulation can run but also the less stable it becomes. Lastly, boundary conditions can be the defined if the system has some limited size. Examples of such conditions are loops (the left boundary connects with the right boundary), hard walls and voids (everything inside is destroyed).

\subsubsection{An example system: Heat in a rod}
As an example we consider a system of heat spreading throughout a thin rod made of a homogeneous material and go through the steps described above. First, we define a state by splitting the rod up into N finite parts of size $\Delta x$ and assigning a temperature value to each, i.e. $u_1$ until $u_N$. In this case the physical laws dictate that heat spreads according to: \begin{equation}
    \frac{\partial u_j^t}{\partial \Delta t} = \alpha\frac{\partial^2 u_j^t}{\partial u_{j-1}^t\partial u_{j+1}^t},
\end{equation} with constant $\alpha$ which using Taylor expansions can be rewritten as:
\begin{equation}
    \frac{\alpha}{\Delta x^2}(u^t_{j+1} - 2u^t_j + u^t_{j-1})
\end{equation}
Which then gives us the following transition function using Forward Euler:
\begin{equation}
   F(u_j^t)  = u_j^t +\frac{\alpha\Delta t}{\Delta x^2}(u^t_{j+1} - 2u^t_j + u^t_{j-1})
\end{equation}
Now we should replace our $\alpha$ with the correct constant, which in this case would be a combination of the density, heat conductivity and thermal mass of the material. The starting state could now also be determined, for example, the rod could be heated on one side at beginning of the simulation. Lastly, boundary conditions need to be specified as the transition function is otherwise undefined near the edges of the rod as $u_{j-1}$ or $u_{j+1}$ do not exist there. A common approach would be to assign $F(u_1^t) = F(u_N^t) = 0$.
\subsection{Simulations in the Medical Domain}
In the medical domain the advantages of simulation (possibly in addition to real experiments) are apparent. In this domain it is often not possible to perform experiments directly due to medical and ethical concerns. Furthermore, collecting high quality scans to study the dynamics of an ongoing procedure is difficult if that procedure must be completed quickly (such as with ischemic stroke). For example, it is difficult to do so when studying how a stent retriever unfolds in a blood clot, as that likely requires pausing the mechanical thrombectomy procedure.

Lastly, there are also processes that can be studied outside the body using real world physical models, called \glspl{phantom} (e.g., an organ taken from a donor, or a fake organ that has similar properties to the real one) However, these \glspl{phantom} are often limited in quantity (in case of real organs) or very labor-intensive to produce \cite{groves2023review} (in case of fake organs). In contrast, physical simulations do not have these disadvantages. They allow for endless experimentation and may involve patients only in minimal ways, such as using the patients physical characteristics as parameters (such as blood pressure or brain structure), which often need to be recorded regardless. However, the costs associated with performing accurate numerical simulations can be high, as the more accurate a simulation needs to be the more computing power is required. One other disadvantage specific to the medical domain is the calibration of model parameters to the real world. For this, actual data needs to be collected on factors like blood viscosity, coefficients of frictions, etc. These are not easy to collect, but fortunately they only need to be collected once and some studies have already done so \cite{boodt2021mechanical}.

\subsection{Physical simulations \& AI} \label{simsAI}
Even though the knowledge of the physical laws and enough compute is in theory enough to simulate any physical system, AI can be a useful tool for physical simulations to achieve results otherwise infeasible. For our purposes, it is the most interesting to discuss how machine learning can speed up simulations. However other possible beneficial approaches exist, such as symmetry finding \cite{wang2310discovering} or generating new examples of stable physical systems such as molecules \cite{walters2020applications}. Before discussing how simulation speed can be increased we will explain how a machine learning model can learn a set of equations in the first place.
\subsubsection{Approximating a Transition function using Machine Learning}
In the explicit case it is straightforward how a sufficiently expressive machine learning model can approximate a set of equations, as $F(s_t)$ can be approximated directly by training a model to predict $s_{t+\Delta t}$ from $s_t$ sampled from previously done simulations. In the implicit case where there is no clear $F(s_t)$, this is less obvious, but as it turns out we can in fact use the exact same approach. This is because by the implicit value theorem, for any set of equations (that follow certain smoothness and differentiability constraints) there exists some function that directly computes the variables (the next state in this case). This function often does not have an analytical solution, but since our machine learning model is capable of approximating any continuous function by the universal approximation theorem this does not pose an issue in practice. Thus, by training to predict $s_{t+\Delta t}$ from $s_t$ we are approximating a function $F'(s_t)$ which is implicitly defined by the set of equations governing the system.

\subsubsection{Speed and Robustness Advantages of Machine Learning}
As we have seen before, for numerical simulations there are tradeoffs between computational speed and accuracy. However, in many cases it is possible to do slow and accurate simulations in advance, which can be taken advantage of. This can be done by using machine learning, which can approximate the $F(s_t)$ function by using these accurate simulations as training data ahead of time. Then when speed is needed the machine learning models only do inference. This comes with a large speedup as in general machine learning methods are much faster than numerical simulations. Since they do not suffer as much from the need for high resolutions and small timesteps for stable results. Moreover, many machine learning methods are highly optimized for speed and can perform most of their calculations in parallel on the GPU, something that is still not widely available for numerical physics simulations.

These machine learning models can thus provide a faster way of emulating these physical processes after training is done, for example in our case a 12 hour thrombectomy simulation \cite{daei2023computational} would be too slow for clinical purposes, but a neural network trained on those simulations could be fast enough to be practical. Papers that have tried this in similar domains have already achieved a 10-1000 times speed up with minimal loss in accuracy \cite{liang2018deep,regazzoni2024learning,wu2022learning}. The fact that the speedup factors take values is such a large range indicates that some tasks are much more difficult emulate than others.

Another benefit of using machine learning models is that they are also more suited to dealing with noisy input data, such as scans of a patients vascular system. This is in contrast to numerical physics simulations, for which great care must be taken to make sure the starting \glspl{mathm} are of the correct resolution and contain no mistakes in their topology. Machine learning methods can be trained for robustness, by incorporating such imperfections in the training data.

\subsubsection{Challenges of Mesh Data}
As we discussed, discretization using meshes has advantages for simulation accuracy, however it also has its challenges when it comes to building a machine learning \gls{numsurr}. Firstly, the amount of nodes used in some simulations is very large, 10.000-100.000+, each of which can possibly interact with every other node. This means that any machine learning method must either be able to handle such large inputs in terms of speed and memory efficiency, or it must be able to work with a sparser down-sampled version. The former can be a major issue for some model architectures like the transformer, which can have its memory cost scale quadratically with the input size. The latter is something that machine learning can deal with as discussed before, but it always comes at some loss of accuracy. Secondly, the exact structure of the mesh can have a large impact on how difficult training is. Some simulations add more mesh nodes in areas were the numerical solver needs more detail. These changes can also impact the performance of the machine learning model that is trained on that data as this means the nodes have a varying density, which the model must be able to handle.

\section{Existing Machine Learning Methods}\label{existingml}
The methods explained in this section are those that are likely to be the most relevant for our use-case. They are either specifically made for use in clinical settings related to ours or have been designed for approximating numerical physics simulations involving meshes. To facilitate our explanations we will first cover two techniques that are present in many of the methods we will cover.
\subsubsection{The Transformer model}
\begin{figure}
    \centering
    \includegraphics[width=0.4\linewidth]{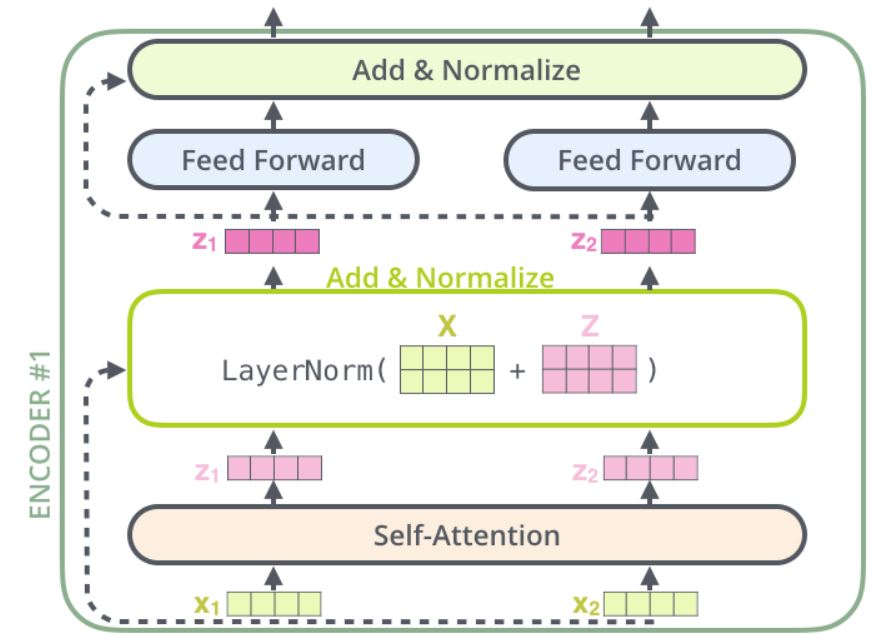}
    \caption{Basic Transformer model. Image shows two inputs. Taken from \cite{Alammar2026}.}
    \label{fig:transform}
\end{figure}

The transformer \cite{vaswani2017attention} is an architecture that uses self-attention in combination with Multi-Layer-Perceptrons (MLP), normalization and skip connections. Its main feature, self-attention, works as follows. For each input of the self-attention block it calculates a query $Q$, a key $K$ and a value $V$, each represented as a vector and each calculated using a different linear projection. Each input is then compared to every other input (including itself) in the following way. First, for input i, the value $Q_i^TK_j$ is calculated for every $j$. The softmax is then taken over these values resulting in attention weights that add to one, these weights represent the relevancy of input $j$ to the output of input $i$. Lastly the attention weights are multiplied by their respective values, i.e. $V$ and summed, resulting in the output for $i$. Putting these steps together gives us the following for output $Z$: \begin{equation}
    Z_i = \sum_j\text{softmax}(Q_i^TK_j)V_j
\end{equation} or in matrix form, where $Q$, $K$, $V$ and $Z$ are stacks of horizontal vectors:
\begin{equation}
    Z = \text{softmax}(QK^T)V
\end{equation}
This then gives us an embedding for each input of the self-attention layer. In most applications, self-attention is done multiple times in parallel, called multi-head self-attention. Each head has its own linear weights for calculating the queries, keys and values and performs self-attention separately. The outputs of each of these heads is then combined with an MLP.

Incorporating attention, the entire structure of a Transformer block then looks like the following, consisting of two parts. The first part starts with multi-head self-attention on the input of the block. Then, using a skip-connection, the input of the transformer block is added to each of these values. Finally a layernorm is applied. The second part takes the output of the first part and applies an MLP to each value separately. Through a skip connection the outputs of the MLP are then connected to the inputs of the second part. Finally layernorm is applied once again. For an illustration see figure \ref{fig:transform}.

\subsubsection{Message Passing Neural Network}
The Message Passing Neural Network (MPNN) \cite{gilmer2017neural} is an architecture designed to specifically work on graphs. To use an MPNN, a graph must be defined by a list of nodes and a list of edges between these nodes. Each node and each edge has a feature vector $v_i$ for node $i$ and $e_{ij}$ for a (possibly directed) edge between node $i$ and $j$. For both the nodes and the edges some starting value must be assigned to their feature vector, which can come directly from the data, or be replaced by some arbitrary value if the data does not contain them.

An MPNN works using a series of message passing blocks, which function as follows. A message passing block works in three steps. In the first step, which is optional, the edges are updated using an MLP $f^E$:
\begin{equation}
      e'_{ij} = f^{E}(e_{ij},v_i,v_j).
\end{equation}
In the second step messages are calculated, using a different MLP $f^M$:
\begin{equation}
    m_{ij} = f^M(v_i,v_j,e'_{ij}).
\end{equation}
Finally, for each node the messages to each node are aggregated and its information is updated using a third MLP $f^V$ and an aggregation function $\sigma$:
\begin{equation}
    v'_i = f^V(v_i,\sigma_j^J(m_{ij})).
\end{equation}
With the sum commonly used as an aggregation function. As can be seen, each message passing layer the information of a node is spread to its neighbors, in other words it increases the receptive field size by 1 for each layer. At the final layer, feature vectors can then be used for downstream tasks, such as edge prediction, node classification or graph classification (by aggregating information from all nodes).
\subsection{End-to-end Approaches}
Before we dive into the methods that can provide a transition function based on simulations we must first discuss some that do not. These methods take an end-to-end approach and take only the geometry at $t=0$ and from that directly predict the physical or clinical outcome. Unfortunately, this end-to-end approach results in a loss of insight into the dynamics of every step between the beginning and the end. At most, these methods can offer insight into the aspects of the input that were relevant to the predicted outcome. For example, such a method could predict that a blood clot will fracture and highlight what part of the geometry it used to make that prediction. While useful, not being able to see how exactly the clot fractured will make it more difficult to adjust the operation such that the fracture would not occur.
 However, while less insightful, end-to-end methods also have the major advantage of having a much simpler task to train on. If such models are accurate and general enough, their increased performance could outweigh their downside. 

\subsubsection{Handcrafted Geometric Features}
Various quantifiable geometric factors have been discovered that have a simple impact on the \gls{physout} of mechanical thrombectomy, such as the length of the clot, how much of the blood vessel is blocked by the clot, the total volume of the clot and the general area of the brains blood vessel system the clot is located \cite{yeo2019does}. These relations however do not account for the fact that the brain's vascular system varies greatly among people and as such have only a small amount of predictive power.

Some approaches take it one step further and try to select geometric features that are more granular. In \cite{daei2023computational} the author performed very detailed simulations that involved stent retrievers removing blood clots from vessels with realistic geometries, each with slightly different starting variables. These simulations were then placed into 4 classes based on their likely \gls{clinout} according to the TIMI grade (see Table 1 in \cite{TIMI1985}). For each simulation, the starting position, clot type and clot length were recorded. These were then used for fitting various ML models to predict the TIMI grade, namely a small MLP, Naïve Bayes, Descision Trees and Random Forests. The results of these methods were however not good enough to be clinically useful, reaching accuracies of around 70 percent.

\subsubsection{Principal Component Analysis and Neural Networks}
In \cite{liang2018deep} the authors use simulations \cite{liang2017machine} of the stress on a blood vessel, the aorta, to train a system of neural networks. In these simulations the blood vessels were divided up into small elements. The pressure inside the blood vessel was increased until at least one element of the blood vessel reached a critical stress value, which in the real world would result in a rupture. A snapshot of this final step was then taken as a datapoint, with the shape of the blood vessel as the input and the stresses on each element as the desired output. This process was then done for each digital aorta model that was available.
This dataset was then used to train a machine learning system that predicts the output given the input using three subsystems. The first compresses information about the shape using principal component analysis combined with a linear layer, the second uses a 3 layer neural network to map these values to a set number of stress values and the third which decodes these stress values using one transposed convolutional layer. These subsystems are all trained individually and then combined into an entire end to end model. 

With this method the authors achieved results that came very close to the simulated results, reaching less than $1\%$ deviation from the numerical physics simulation when measuring the Von Mises stress (a measure of stress that combines different types of stress). The speed up of their method was around $1800$ times compared to the full numerical physics simulation. 


\subsection{Transolver}

\begin{figure}[h]
    \centering
    \includegraphics[width=\linewidth]{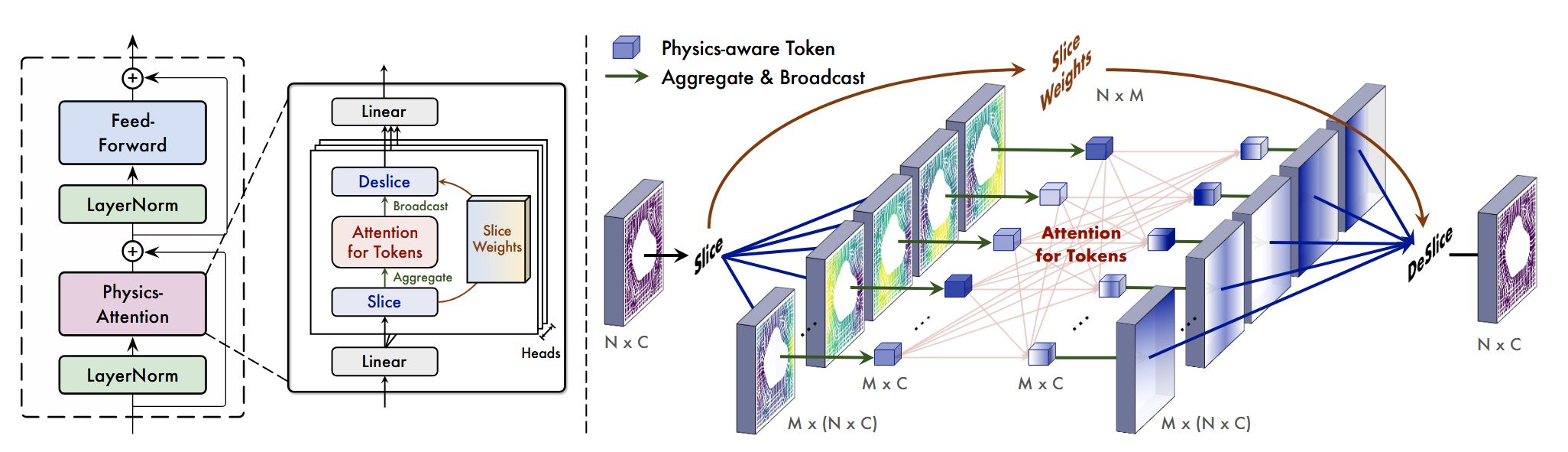}
    \caption{Overview of the Transolver architecture. On the left the Transolver blocks can be seen, which is a modified version of the transformer architecture. On the right the PhysicsAttention mechanism is shown in more detail for a single attention head. Taken from: \cite{wu2402transolver}.}
    \label{fig:blood clots}
\end{figure}
The Transolver \cite{wu2402transolver} is a method that predicts features of each input node and is based on the transformer architecture. Its core idea is to split the input nodes into slices based on physical similarity and then performing attention between slices. This allows for the attention mechanism to scale with the amount of slices instead of being performed between every node.

The Transolver method on meshes works as follows. First each nodes information, a concatenation of its geometric information and its other properties (for example velocity, pressure etc), is processed with a small MLP. Then the nodes go through a set of Transolver blocks, which special feature is the Physics Attention block which replaces normal attention. 

These Physics attention blocks work as follows. At the start of each of these blocks each input node is put through two separate linear layers, the first computes weights using a softmax operation, which indicates how much a node belongs to each slice of each head. This results in weights $w_{ijk}$ for each input $i \in N$, attention head $j \in H$ and slice $k \in M$, where $\sum^M_m w_{ijm} = 1 $ for all $i$ and $j$. The second linear layer simply projects an input node to the embedding dimensions of each slice of each head. For each slice and each head all nodes are then aggregated into one feature using a normalized weighted average in the following way: \begin{equation}
    z_{jk} = \frac{\sum^N_nw_{njk}x_n}{\sum^N_nw_{njk}}
\end{equation}For each head attention is done  on this aggregated information between all slices, the output of this we will call $z'_{jk}$. As a second to last step of the Physics Attention block, for each head the nodes are reconstructed using a deslicing operation: \begin{equation}
    x'_{ij} = \sum^M_mw_{ijm}z'_{jm}
\end{equation}Finally, following the multi-head attention paradigm, the outputs for each node are aggregated across attention heads using a linear layer. 

These Transolver blocks then consist of a LayerNorm operation, a Physics Attention block, another Layernorm operation and finally a small MLP. Using the same structure of LayerNorms and skip connections as the normal Transformer. After applying a number Transolver blocks a linear layer projects the information of each node to the correct output dimension.


\subsection{Erwin}
\begin{figure}[h]
    \centering
    \includegraphics[width=1.0\linewidth]{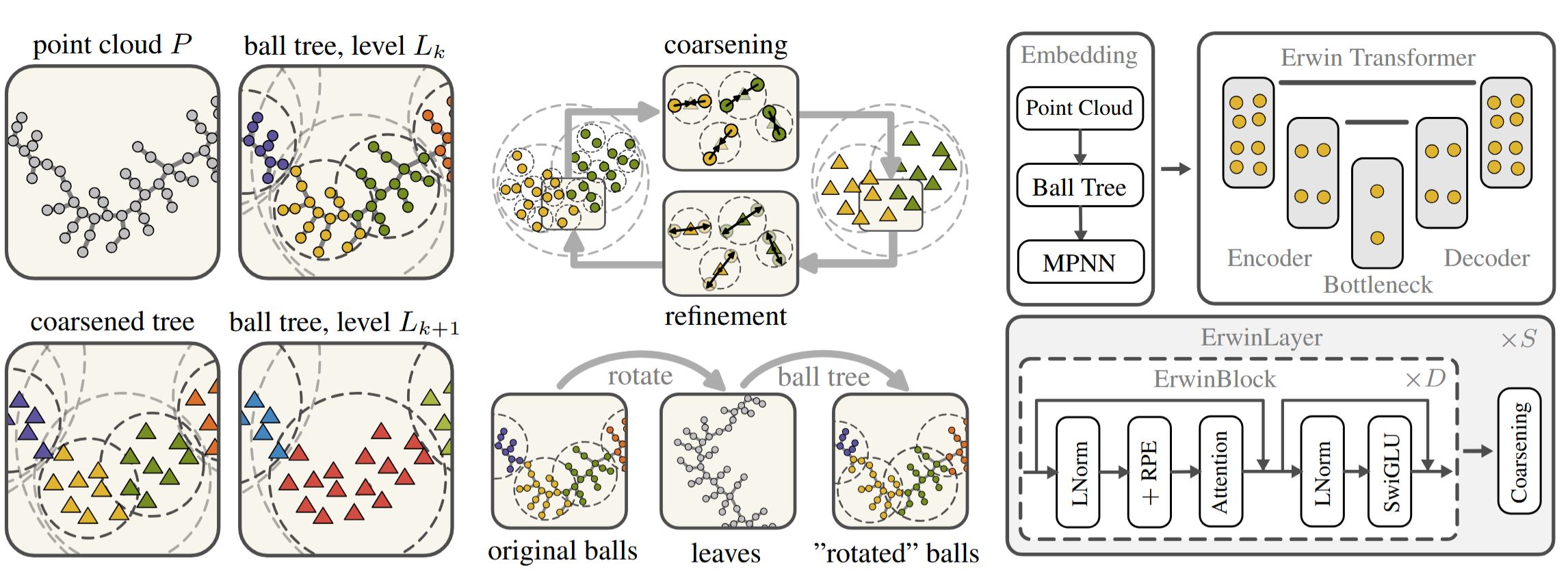}
    \caption{Overview of the Erwin Architecture. On the left the construction of the nodes into a balltree can be seen. In the top image in the middle, an example coarsening and refinement operation is displayed. The image on the bottom in the middle shows how the rotation changes which ball contains which nodes when constructing the ball tree. In the top right the general U-shape of the architecture can be seen. In the bottom right a BallProcessing block is displayed (referred to as an ErwinLayer in the image). Taken from: \cite{zhdanov2025erwin}}
    \label{fig:placeholder}
\end{figure}
Erwin \cite{zhdanov2025erwin} is a hierarchical transformer based model that performs attention within subsets of the input, combines those subsets and then repeats this process. These subsets take the form of balls, which have a center and a radius and which divide the nodes into disjoint subsets spatially. It decides which nodes belong to which ball at each level by constructing a ball-tree, which recursively divides the nodes between balls of equal size. At the first level self-attention is performed within balls of input points and the output of this results in a singular feature vector for each ball. On subsequent levels these singular feature vectors are then divided into balls again (using the centers of the previous level as positions) and the process repeats. Erwin employs some additional tricks to improve performance. Firstly, it can do additional coarsening, which can combine multiple feature vectors using a linear projection, resulting in a bigger receptive field. Secondly, it constructs the ball-tree in an efficient way such that it forms a perfect binary tree which allows for fast indexing. Thirdly, which nodes fall into which ball is slightly different every other layer which is achieved by rotation, resulting in more cross-ball information exchange. Lastly, it uses a learned embedding in the form of a Message Passing Neural Network (MPNN), see section \ref{existingml}.

In detail, Erwin works as follows. First, we will explain the BallProcessing block as it is used throughout the architecture. In each BallProcessing block for each ball a positional encoding is applied according to the following formula: \begin{equation}
    X_{Bi} = X_{Bi} + (P_{Bi} + c_B) W_{pos}
\end{equation}where $X_{Bi}$ is the feature vector of node $i$ in ball $B$ and $P_{Bi}$ its position, $C_B$ the center of the ball and $W_{pos}$ a learnable projection. Then a number of BallAttention blocks is applied, which compute mulit-head self-attention and then use a Swiglu MLP \cite{dauphin2017language} \cite{shazeer2020glu} using Root Mean Square Normalization \cite{zhang2019root} and skip connections.  Additionally these BallAttention blocks use a distance fall-off that is applied to the attention values according to: \begin{equation}
    D_{Bij} = \sigma^2||P_{Bi} -P_{Bj}||_2
\end{equation}with $\sigma$ a learnable parameter and $i \not = j$. After the BallAttention blocks an optional coarsening or refinement operation is applied. The coarsening operation selects an amount of nodes equal to the stride size in the form of a ball, concatenates their features and projects them resulting in a decrease in the amount of nodes and an increase in receptive field size. It does this according to: \begin{equation}
    x_{B'} = \text{concat}_{i\in B}(\text{conca}t(X_{Bi},P_{Bi} - c_B)) W_c
\end{equation}
The refinement operation works by projecting a singular point back up to all the points in a ball as such: For all nodes $i\in B'$ do \begin{equation}
    X_{B'i} = \text{concat}(X_{Bi},P_{Bi})W_r
\end{equation}
\begin{figure}
    \centering
    \includegraphics[width=0.5\linewidth]{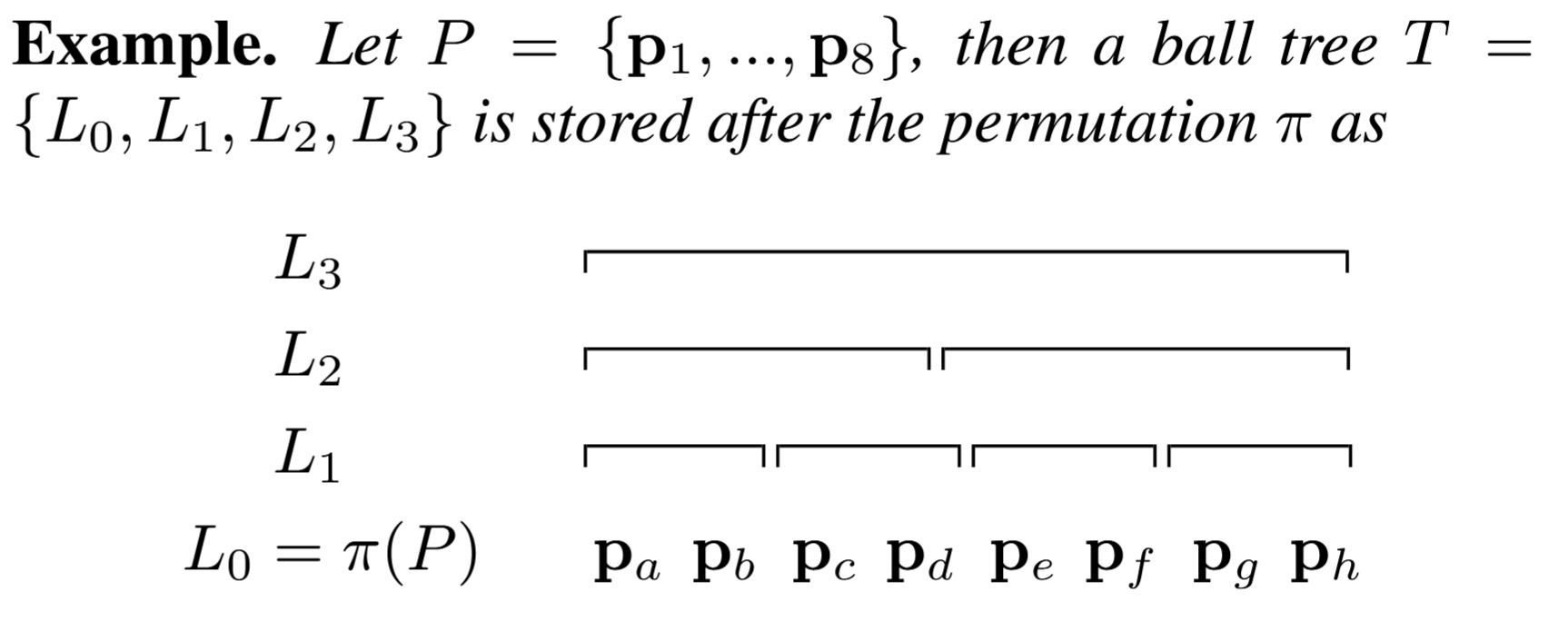}
    \caption{Example balltree formation. $P$ is the set of nodes and $T$ is the collection of layers of the tree. The permutation $\pi()$ could for example be a rotation as described below. Taken directly from \cite{zhdanov2025erwin}.}
    \label{fig:placeholder}
\end{figure}
The BallProcessing blocks are made more efficient by the underlying binary tree the ball structure is stored in. The construction of this tree is done recursively starting from the complete set of nodes. At each step the current level is split into two sets of identical size, introducing virtual nodes when necessary to keep the sizes the same. This split is made along the axis (x, y or z) with the largest difference between the highest and the lowest value and is split according to the median of that axis. Doing this allows for two things, firstly it allows for fast indexing by reshaping the nodes using tensor operations. For example, reshaping a tensor of $32$ nodes into a $[8,4]$ tensor gives eight balls of size $4$. Secondly, since the ball tree is constructed according to the axis it is possible to construct a distinct ball tree by rotating the input $45$ degrees. This results in the nodes falling into different balls, altering which nodes interact with the attention mechanism. In the architecture the alternative ball tree is used every other BallProcessing block, to increase cross-ball connections.

The complete architecture then looks like the following. Firstly, it embeds the nodes with a MPNN with mean pooling using the provided node features and the pairwise distances as edge features. Secondly, the main encoder is applied, which consists of BallProcessing blocks that each use the coarsening operation. Thirdly, a bottleneck is applied, this is a single BallProcessing block, without any coarsening or refinement. Lasty, the decoder is applied, similarly to the encoder but now with refinement operations. Crucially, the encoder and the decoder are connected with skip connections in a U-net \cite{ronneberger2015u} like fashion (although with addition instead of the concatenation of U-net), with each coarsened level of the encoder connected to its corresponding level of the decoder. As such the amount of BallProcessing blocks in each must be the same. 

\subsection{MeshGraphNet}

\begin{figure}[h]
    \centering
    \includegraphics[width=1.0\linewidth]{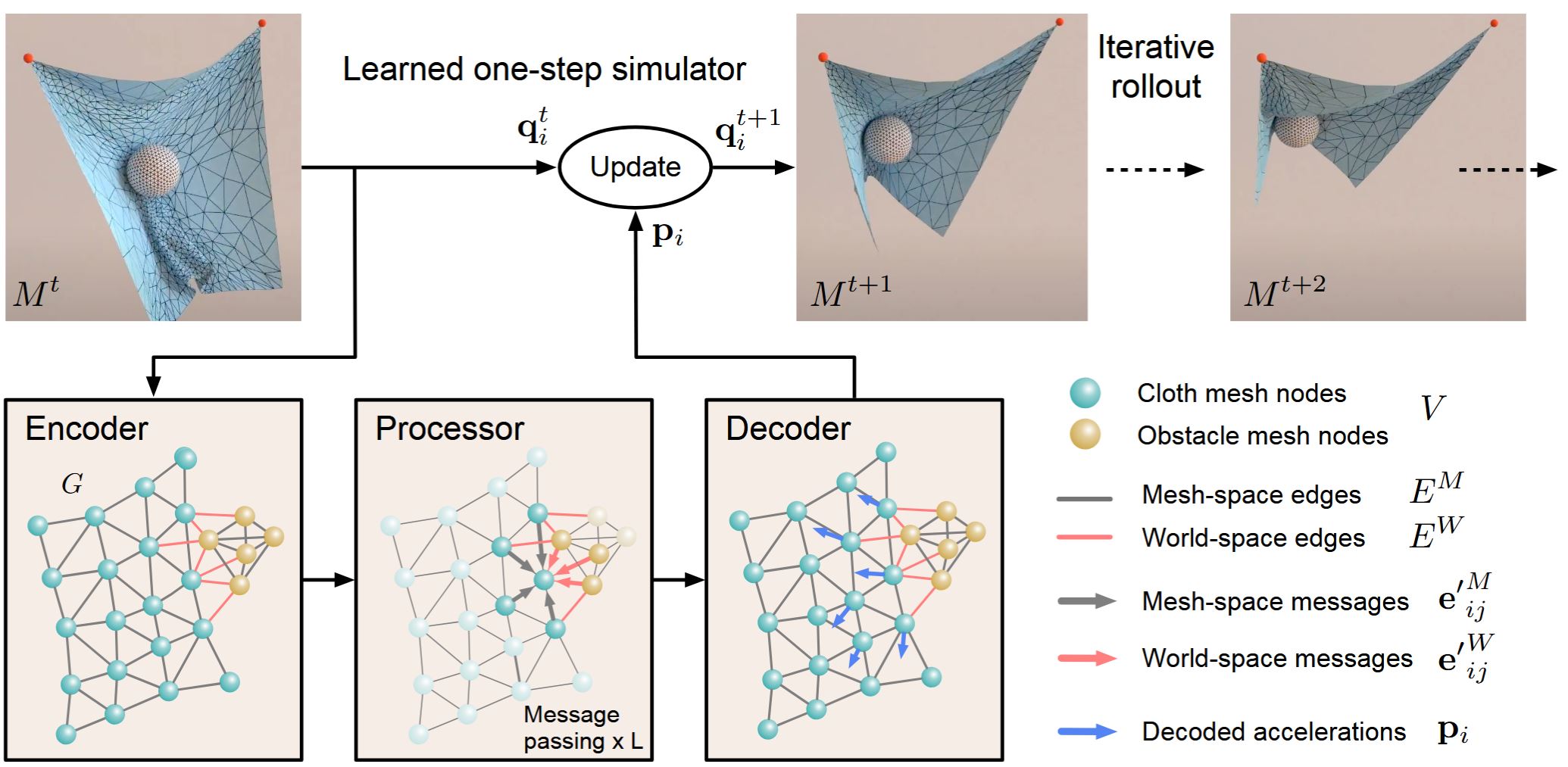}
    \caption{Overview of the architecture of MeshGraphNet. As can be seen it uses an encoder, followed my message passing layer and a decoder. Taken from: \cite{pfaff2020learning}}.
    \label{fig:placeholder}
\end{figure}
MeshGraphNet \cite{pfaff2020learning} is a graph neural network \cite{gori2005new} that introduces some additional techniques, like processing mesh and world edges separately, to work well on \glspl{mathm}. 
  MeshGraphNet then works as follows. Firstly, it encodes the edge and node information separately using two MLP's, which gives an embedding for every edge i.e. $e_{ij}$ for an an edge between node $i$ and $j$ and an embedding for every node i.e. $v_i$ for node $i$. Secondly, it does a set of message passing layers, as explained in 
  section \ref{existingml}, each of which uses three separate MLP's, two for the edges and one for the nodes. In the paper, they describe that they differentiate between mesh edges, which represent a physical link between two nodes, and world edges, which represent that two nodes may interact due to physical proximity. This then gives us the following, where $f^{EM}$ and $f^{EW}$ are the edge MLP's for mesh edges and world edges respectively and $f^N$ is the node MLP, both using a residual connection: 
  \begin{equation}
      e'^{M}_{ij} = f^{EM}(e_{ij}^M,v_i,v_j),
  \end{equation} 
  \begin{equation}
      e'^{W}_{ij} = f^{EW}(e_{ij}^W,v_i,v_j),
  \end{equation} 
  \begin{equation}
      v'_i = f^N(v_i,\sum_je'_{ij}).
  \end{equation} As can be seen it uses the sum operator to aggregate the edge information when updating the nodes, summing over both world and mesh edges. Furthermore, no explicit message is calculated. Instead the newly calculated edge feature vector is simply used instead. Lastly, an MLP is used to decode the node features to the relevant number of output dimensions.

In addition to this architecture, MeshGraphNet uses two additional techniques. Firstly, it creates edges between nodes that are physically close, i.e. if $|x_i - x_j| < r$ for some range $r$ so as to model interactions between nodes that might be far removed from each other in the mesh. Secondly, at every timestep it recalculates where on the surface of the object the resolution needs to be high and where it does not. It then adds and removes nodes in certain areas where needed. It does this by using a domain specific remesher during training and learning parameters for a generic remesher from it for use during inference. Importantly, this technique is completely depended on the availability of such a remesher specific to the dataset and can therefore not always be used.


\section{Bary Centric Coordinate Sampling}
In this section we describe a method for sampling new pointcloud datapoints from existing pointclouds that evolve over time, provided that the order and the amount of nodes is consistent. The main idea of the method is to pick a triangle and sample a point on it in such a way that the relative position of that point on the triangle stays the same across timesteps. This then allows a machine learning model to predict the next location of the point, as it moves consistently and with similar dynamics as the nodes that form the triangle. 

To sample a point on a triangle so called barycentric coordinates are used, which relate the point to the three vertices of the triangle. Specifically, with barycentric coordinates $\lambda_1$, $\lambda_2$ and $\lambda_3$ the position of a point $p_s$ in the Cartesian coordinate system becomes \begin{equation}
    p_s = \lambda_1p_1 + \lambda_2 p_2+ \lambda_3 p_3
\end{equation}
with $p_1$ to $p_3$ representing the coordinates of the vertices of the triangle and with all positions represented as vectors. If we add the constraint that for each barycentric coordinate $\lambda \in [0,1]$ and that they sum to $1$ then our barycentric coordinates define a unique point that lays on the triangle.

We can then sample barycentric coordinates with respect to this constraint to sample a point on the triangle. If we then keep the barycentric coordinates stationary across time, in the Cartesian coordinate system, it moves on the triangle proportionally to the movement of the vertices, as desired. 

To use this approach to create a new datapoint we must sample points on many triangles across two timesteps, the first step as input for the model and the second to calculate the loss. We provide the pseudocode in algorithm 1.
\begin{algorithm}[H]

\caption{BaryCentricSampling}
\begin{algorithmic}[1]

\Require\\
$\mathcal{N}_1$ : set of node positions at timestep 1 \\
$\mathcal{N}_2$ : set of node positions at timestep 2 \\
$\mathcal{E}$ : set of edges between nodes in $\mathcal{N}_1$ \Comment{Ex. Constructed based on distance}\\
$n$ : amount of nodes to sample 
\Ensure\\
$\mathcal{P}_1$ : sampled point cloud at timestep 1 \\
$\mathcal{P}_2$ : sampled point cloud at timestep 2

\State $\mathcal{T} \gets \textsc{FindTriangles}(\mathcal{E})$
\State $\mathcal{P}_1 \gets [\ ]$
\State $\mathcal{P}_2 \gets [\ ]$

\For{$1$ to $n$}  
    \State Sample $(c_1, c_2, c_3)$ from $[0,1]$ such that
    $c_1 + c_2 + c_3 = 1$ 
    \State Sample triangle $t \in \mathcal{T}$
    \State Let $(v_1, v_2, v_3)$ be the vertices of $t$
    \State $\mathbf{p}_1 \gets c_1 \mathcal{N}_1[v_1] + c_2 \mathcal{N}_1[v_2] + c_3 \mathcal{N}_1[v_3]$
    \State $\mathbf{p}_2 \gets c_1 \mathcal{N}_2[v_1] + c_2 \mathcal{N}_2[v_2] + c_3 \mathcal{N}_2[v_3]$
    \State Append $\mathbf{p}_1$ to $\mathcal{P}_1$
    \State Append $\mathbf{p}_2$ to $\mathcal{P}_2$
\EndFor

\Return $\mathcal{P}_1, \mathcal{P}_2$
\end{algorithmic}
\end{algorithm}
\chapter{Method}
This chapter describes our experiments to evaluate machine learning based step-by-step surrogate models for numerical mechanical thrombectomy simulations. First, we describe the datasets used, including one dataset with simple and one with complex geometry. Second, we describe how we configured Erwin, Transolver and MeshGraphNet to for these tasks, followed by the training procedure used. Finally, we present the experiments that will answer our research questions.
\section{Datasets}
The datasets presented here were build by inSteps B.V. using the implicit finite element solver Abaqus from Dassault Systèmes. The hardware this was done with was an AMD Ryzen 7900 X CPU using 4 cores. The simulations were finetuned and checked by hand to ensure their stability and to ensure the blood clots were of the appropriate size to get stuck in one of the blood vessels. These simulations were then turned into datapoints such that the input of each datapoint consists of the current node positions, the previous node positions and a one hot encoding of the node types (representing either blood clot or blood vessel), with the task being to predict the node positions in the next timestep.
For both datasets simulations were done from multiple initial conditions to increase the variation of the data. After simulation, each dataset was scaled to $[-1,1]$ to provide a consistent scale.
\begin{figure}[h]
    \centering
    \begin{subfigure}{0.48\linewidth}
        \centering
        \includegraphics[width=\linewidth]{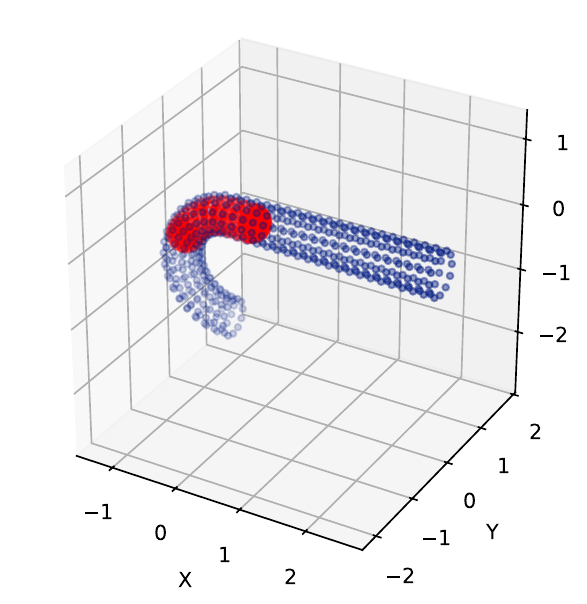}
        \caption{}
        \label{fig:bend_vessel}
    \end{subfigure}
    \hfill
    \begin{subfigure}{0.48\linewidth}
        \centering
        \includegraphics[width=\linewidth]{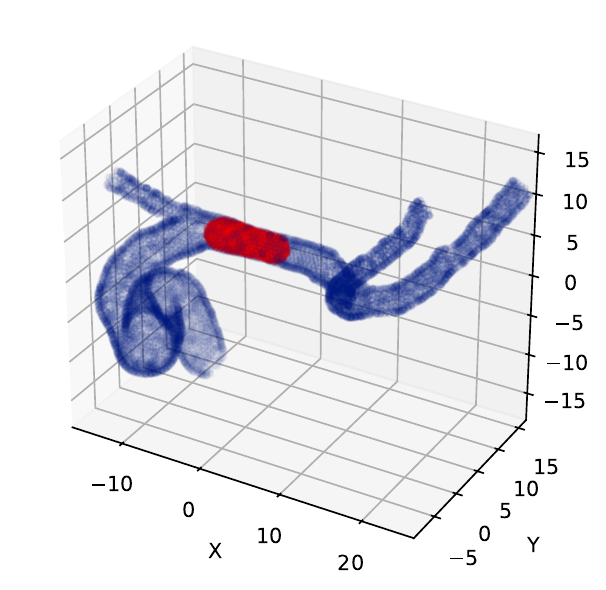}
        \caption{}
        \label{fig:clot_entry}
    \end{subfigure}
    \caption{Frame from BendVessel (left) and ClotEntry (right). The blood clot is shown in red and the vessel walls in blue. Both frames show a blood clot stuck in the blood vessel.}
    \label{fig:side_by_side}
\end{figure}

\subsection{BendVessel}
 BendVessel consists of a simplified blood vessel with a bend in it. In this bend a blood clot is placed and pushed out by applying a force, to emulate an aspiration device pulling it out. For this dataset, 20 different simulations were run each slightly rotated (from 0-180 degrees around one axis) and with slightly different initial conditions. Each simulation contains 977 nodes and each simulation lasts 40 timesteps. This dataset resembles a smaller data regime as it only has 800 datapoints and each datapoint has a small number of nodes.
 In our datasets, we consider an 'average' blood clot with a homogeneous composition throughout and of an average size. The average simulation time for this dataset was 3 minutes.

\subsection{ClotEntry}
ClotEntry consists of 10 realistic synthetic blood vessels, taken from \cite{kuipers2024generating} which generated these using a machine learning model that was trained on real blood vessels. In each of these vessels a blood clot is placed and a force is applied to it, in the same manner as for BendVessel. Each simulation begins with a stuck blood clot and ends when the blood clot has fully exited the entire blood vessel, which is around 10 times the length of the clot. As such, a large part of the dataset involves the blood clot moving dynamically around the blood vessel. ClotEntry consists of 10 simulations each with 200 timesteps and with around 27000 nodes. This dataset represents an intermediate data regime, as there are 2000 datapoints and each datapoint is large enough to challenge current machine learning architectures when it comes to memory efficiency and speed. The average simulation time for this dataset was 25 minutes.

\section{Machine Learning Models}
Since MeshGraphNet and Erwin require edges between nodes to function we generate edges between nodes at runtime based on pairwise distance as we assume we are not given any (just as we would need to do when simply given a scan of a patients vascular system). We do this by connecting nodes within a radius $r$, i.e. if $|x_i - x_j| < r$.
For each model a hyperparameter search was done using the Bayesian method, including searching for the optimal $r$. Importantly, this search was done once on the ClotEntry dataset using $\pi$ radians of rotations on the training data and 2000 nodes. These hyperparameters were then used for every experiment. To see the parameters used in the final tests see Appendix \ref{hyper}.
\subsection{Erwin}
Erwin was used as provided by the official Github implementation, with a small MLP on top to project down to the correct dimension. The hyperparameter search was done on a smaller subdomain of the total possibilities of the hyperparameters, as the amount of combinations grows rapidly with the amount of BallProcessing layers. Instead, only the depth (i.e. the amount of BallAttention blocks) and the head count of the first BallProcessing layer of the encoder and of the decoder were directly searched over. The deeper layers of the encoder and decoder were made by simply adding or subtracting a hyperparameter value from the previous layer, massively reducing the search space. For example if the first BallProcessing layer of the encoder had depth 10 and the hyperparameter value was 2 then the second layer would have depth 12, the third depth 14 and so on. Attention head count was kept the same for each BallProcessing layer (for the encoder and the decoder separately).
\subsection{Transolver}
The main part of the Transolver model was taken from the Nvidea PhysicsNemo library \cite{physicsnemo2023}. By default, Transolver requires a spatial embedding, but the paper does not provide any guidance on how to define and calculate it. To provide one, we used the same MPNN based embedding as Erwin. Using this learned embedding ensures our results were not skewed by the selection of the wrong spatial features (which must be picked by hand otherwise).

\subsection{MeshGraphNet}
The complete MeshGraphNet model was taken from the Nvidea PhysicsNemo library \cite{physicsnemo2023}. Just as in the paper \cite{pfaff2020learning} we use edge information consisting of two parts: the vector between the nodes i.e. $x_i - x_j$ if there is an edge between node $i$ and node $j$, where $x$ represents the 3D position of a node and the norm of that vector i.e. $|x_i - x_j|$. It must be noted that in the code of the MeshGraphNet paper and that of the NVIDIA implementation the world and mesh edges are treated as the same, are concatenated and use a single MLP. Here, we will also do so for consistency. To adapt MeshGraphNet to our datasets, we gave our point cloud data in the form of a graph as input to the model using the edges we constructed. 

\section{Training Parameters}
For each dataset we train each model on transitions from $s_t$ to $s_{t+\Delta t}$ from a subset of the simulations, leaving all transitions from one simulation as a test set (and two others for validation purposes). Training was done using an A100 Nvidea GPU with 40 GB of VRAM. This slightly older hardware was chosen to provide a scenario that is more realistic to the types of hardware hospitals might have access to. Training is done using the AdamW optimizer \cite{loshchilov2017decoupled} using a learning rate determined by the hyperparameter search for 300 epochs. This learning rate is scaled according to a warmup of 10 epochs, followed by cosine decay, see figure \ref{fig:learningrate} for a visualization. 

Before predicting the next timestep, an approximation of the node velocities was calculated with $v_t = s_t - s_{t-\Delta t}$, which was then given to the model in addition to $s_t$. Models were evaluated according to the Mean Square Error (MSE): \begin{equation}
    \text{MSE} = \frac{1}{3|C|}\sum^C_{i}\sum^{\{x,y,z \}}_j((\hat{s}_{t+\Delta t})_{ij} - (s_{t+\Delta t})_{ij})^2
\end{equation} where $\hat{s}_{t+\Delta t}$ is the predicted next step, $C$ is the list of indexes of blood clot points and the state $s$ consists of a matrix of stacked vectors of the node positions. As can be seen, the loss is masked so that only the loss that comes from predicting the blood clot is counted, as in our simulations the vessel walls are assumed to be rigid (as discussed in section \ref{bloodvessel}). 

For ClotEntry, we sampled a subset of size 2048 out of the available 27000 nodes each time we trained on a sample. This allows for much faster inference and training, while still giving an insightful visual result. For BendVessel, we used all available nodes.
\section{Experiments}
\subsection{Data Augmentations}
To investigate the impact of rotational data augmentations, datapoints from both datasets were randomly rotated in all 3 dimensions up to a maximum angle. For each model, we measured the loss for a range of maximum angle values. For BendVessel, we used BaryCentricSampling and measured the loss for different amounts of generated data. For each of these experiments, we used the optimal values found for further experiments, to facilitate this we evaluate our models on the validation set. Furthermore, we also performed an experiment on the ClotEntry dataset where we changed the amount of sampled nodes to a range of $[500-6000]$ to investigate how well each of our models scales with the node count.
\subsection{Performance on the Test Set and multi-step rollout}
Finally, we combined all our best data augmentations and compared the single-step performance of our three models over 5 different seeds. To evaluate performance we showed the loss on the test set when predicting the next timestep. Additionally, we perform multi-timestep rollouts of a single seed on the training data to show reliability in longer time domains. To make these rollouts more reliable, the first steps of each simulation were not used as they involve the blood clot being partially outside the entrance to the blood vessel. We cut the first 40 timesteps for ClotEntry and the first 5 for BendVessel. Additionally, during rollout we collect inference speed data.
\subsection{Generalization on unseen Geometries}
  To gain a deeper understanding of the impact of rotational augmentations we applied rotations in a non-random way on the BendVessel dataset, starting with only data points rotated 0 degrees and predicting a datapoint rotated 180 degrees. We then introduced larger and larger rotations until we trained on every datapoint (0 - 170 degrees of rotations) except the 180 degree one to observe the impact on generalizability.

\chapter{Results}

We compared Erwin, Transolver and MeshGraphNet on the BendVessel and the ClotEntry datasets. We first studied the impact of data augmentations on single-step validation loss, namely rotations, datapoint size and data generation. We then used the optimal data augmentation configuration found in those experiments to compare the performance of all three models on the test sets, again for a single step. Additionally, we investigated the loss on a multi-step rollout, while measuring inference speed. Finally, we performed an ablation study to investigate the how well Erwin generalizes to unseen rotations.
\section{Data Augmentation}\label{data_aug}
\begin{figure}[ht]
    \centering
    \begin{subfigure}{0.48\linewidth}
        \centering
        \includegraphics[width=\linewidth]{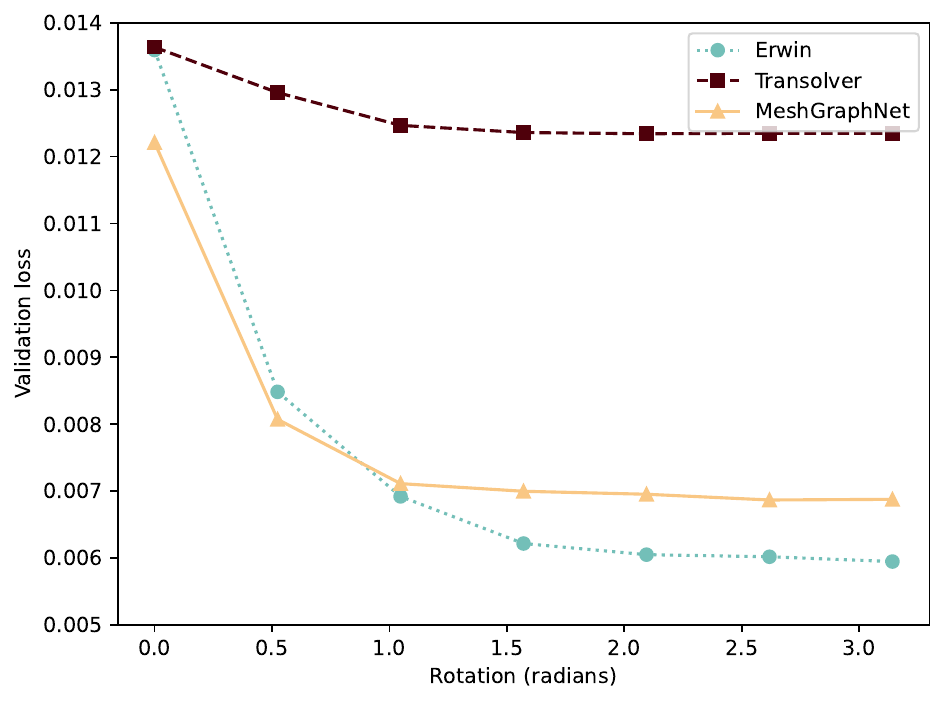}
        \caption{}
        \label{fig:num_nodes}
    \end{subfigure}
    \hfill
    \begin{subfigure}{0.48\linewidth}
        \centering
        \includegraphics[width=\linewidth]{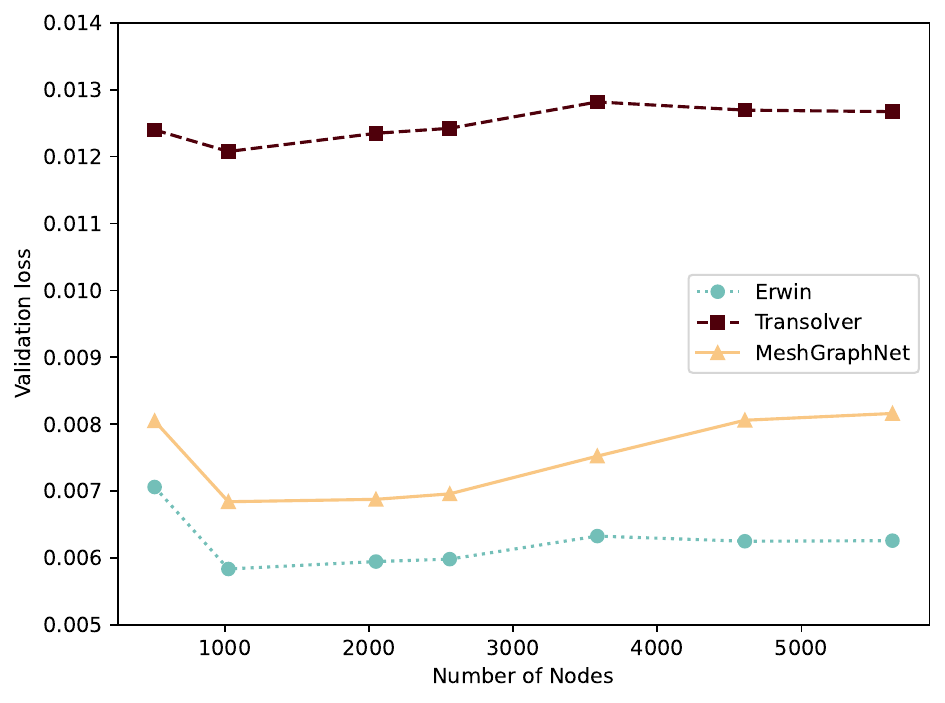}
        \caption{}
        \label{fig:rotation}
    \end{subfigure}
    \caption{Effects on performance of data augmentations for the validation set of the ClotEntry dataset. Figure a shows the impact of random rotations up to the rotation value specified. Figure b shows the impact of sampling different amounts of nodes from the maximum of 27000.}
    \label{fig:side_by_side}
\end{figure}

As can be seen in figure \ref{fig:side_by_side}.a, increasing the rotations of the training set increased performance on ClotEntry across all models, with the increase for Erwin and MeshGraphNet being especially pronounced. In figure \ref{fig:side_by_side}.b, a sharp increase in performance can be seen in the step from 500 to 1000 nodes. However, after this the number of nodes in each datapoint only has a marginal impact on the the performance past 1000 nodes for Erwin and Transolver. In contrast, the loss for MeshGraphNet increases slowly after 2500 nodes has been reached.

\begin{figure}[H]
    \centering
    \includegraphics[width=0.5\linewidth]{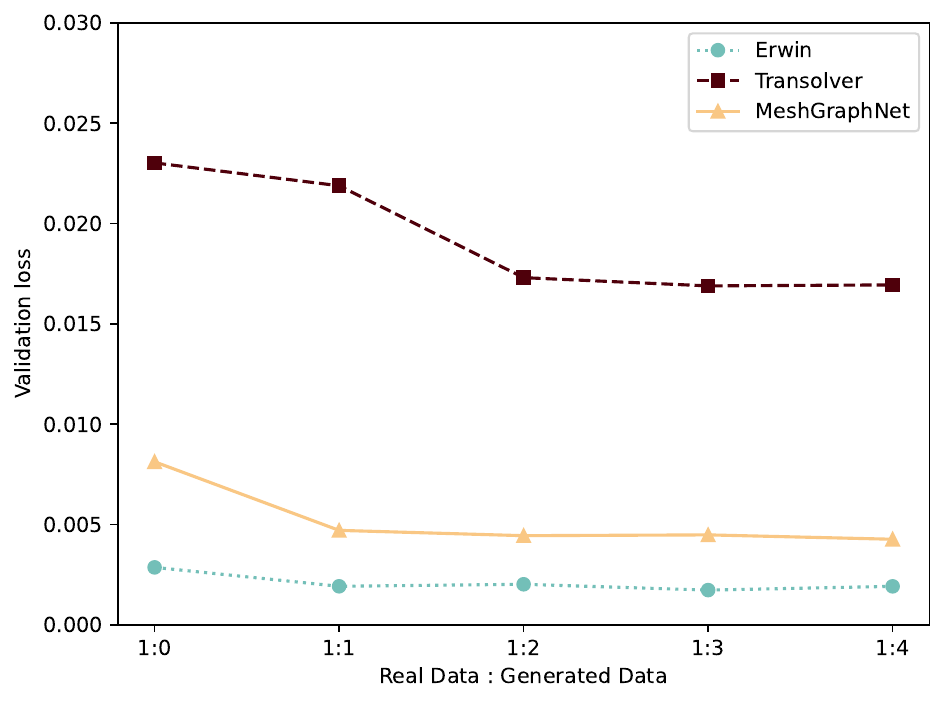}
    \caption{The effect on performance of adding generated datapoints using BaryCentric sampling on the BendVessel dataset for different ratios between real and generated data.}
    \label{fig:bend_gen}
\end{figure}
Figure \ref{fig:bend_gen} shows that performance increases when adding generated data in an amount equal to the real training data. However, this benefit quickly plateaus when the ratio is increased to more than $1:2$ real data to generated data.

\section{Suitability as Surrogate Models}
\begin{table}[h]
\centering
\begin{tabular}{lccc}
\hline
Dataset        & Erwin        & Transolver   & MeshGraphNet \\ 
\hline
BendVessel  & $0.00166 \pm 0.00012$ & $0.02336 \pm 0.01016$ & $0.00469 \pm 0.00052$ \\
ClotEntry   & $0.00799 \pm 0.00044$ & $0.01434 \pm 0.00020$ & $0.00784 \pm 0.00014$ \\
\hline
\end{tabular}
\caption{MSE loss on test set across 5 runs (mean $\pm$ std). Using the best data augmentations found in section \ref{data_aug}, namely $\pi$ radians rotation for both datasets, a 1:2 ratio of real data to generated data for BendVessel and 2000 nodes for ClotEntry.}
\label{main_res}
\end{table}

As shown in table \ref{main_res} Erwin achieves the lowest test loss for BendVessel, followed by MeshGraphNet. On ClotEntry, Erwin and MeshGraphNet perform similarly, within one standard deviation of each other. On both datasets Transolver has the highest loss by a large margin.

\begin{figure}[H]
    \centering
    \includegraphics[width=\linewidth]{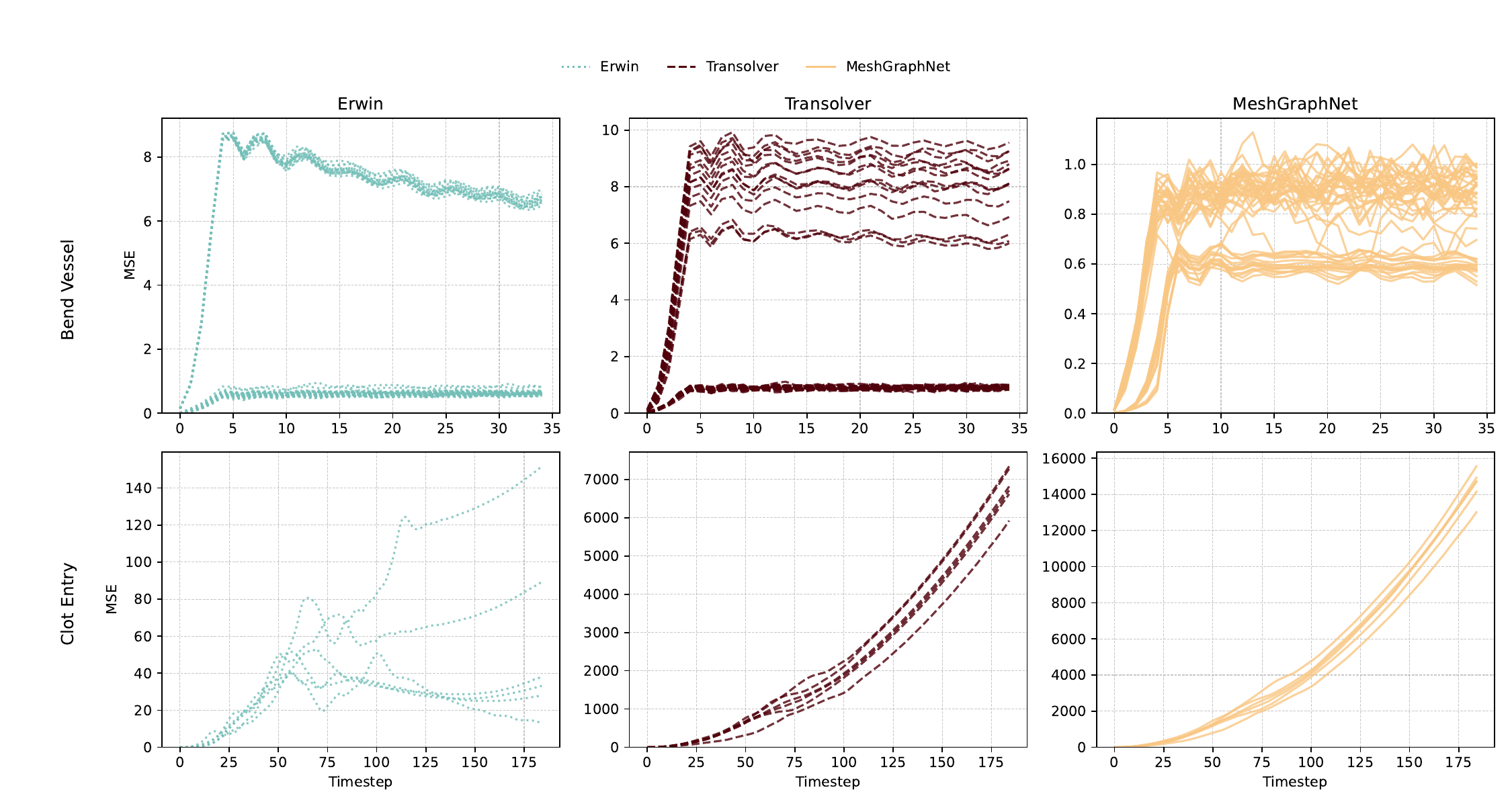}
    \caption{Rollouts for Erwin, Transolver and MeshGraphNet on BendVessel and ClotEntry. Rollouts are performed on all training set simulations. Note that each subplot has a different y-axis scale.}
    \label{fig:rollout}
\end{figure}
As seen in figure \ref{fig:rollout} the performances of our models on multi-step rollout between BendVessel and ClotEntry were quite different.

On BendVessel, the loss of all models stabilized after an initial increase. Each rollout of each model seemed to stabilize into one of two different regions, with the bottom region having a low loss for each model. The top region differed per model, with Erwin and Transolver both having a high loss, while the top region of MeshGraphNet had a loss just a little bit higher than its lower region.

For ClotEntry, the loss of each model diverged rapidly. The values of the loss were much higher than the size of the blood vessel indicating that the blood clot moved outside the blood vessel and away from it.

\begin{table}[H]
\centering
\begin{tabular}{llcccc}
\hline
Dataset & Metric & Erwin & Transolver & MeshGraphNet & Abaqus \\
\hline
\multirow{2}{*}{BendVessel}
& Rollout time &
$2.167 \pm 0.010$ &
$0.65 \pm 0.003$ &
$0.56 \pm 0.003$ &
180 \\
& Speedup &
$\times 82$ &
$\times 278$ &
$\times 322$ &
-- \\
\hline
\multirow{2}{*}{ClotEntry}
& Rollout time &
$13.19 \pm 0.03$ &
$4.33 \pm 0.02$ &
$3.77 \pm 0.02$ &
1500 \\
& Speedup &
$\times 113$ &
$\times 346$ &
$\times 398$ &
-- \\
\hline
\end{tabular}
\caption{Average rollout times for one simulation in seconds (mean $\pm$ std) and speedups relative to the traditional solver (Abaqus). The results shown are multiplied by $\frac{40}{35}$ for BendVessel and $\frac{200}{160}$ for ClotEntry to adjust for the fact that the beginning of each simulation was not included in rollout for the models, but was included for the traditional solver.}
\label{tab:solver_comparison_pm}
\end{table}

The results in table \ref{tab:solver_comparison_pm} show that MeshGraphNet performed the fastest rollouts on both datasets, closely followed by Transolver, while Erwin was significantly slower than the other models. Compared to the traditional solvers all models showed a significant speedup.
\section{Generalization to unseen Geometries}
\begin{figure}[H]
    \centering
    \includegraphics[width=0.5\linewidth]{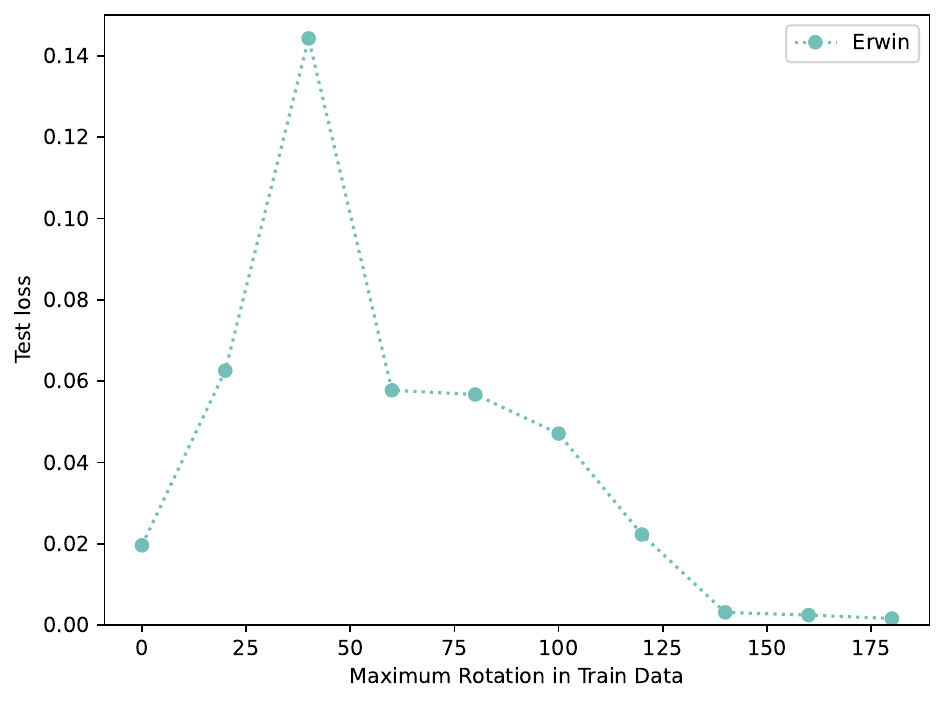}
    \caption{Effect of including different rotations in the training data on performance on the test set for Erwin on BendVessel. Each simulation in BendVessel uses the exact same geometry rotated around one axis in 10 degree increments, with the test set rotated 190 degrees. The x-axis displays the simulations that are included in the training set. For each x-value all simulations are included that have a rotation between 0 and that x-value inclusive (for example, an x-value of 20 entails the inclusion of simulations for 0, 10 and 20 degrees).}
    \label{fig:ablation}
\end{figure}
As shown in figure \ref{fig:ablation} the loss decreases when the maximum rotation of the training data comes close to the rotation of the test data. However, when the maximum rotation is not close to the test data rotation the loss is much higher, spiking at maximum rotation $40$. 

\chapter{Discussion}
In this thesis, we investigated how well current machine learning methods are suited as step-by-step surrogate models for numerical mechanical thrombectomy simulations by training Erwin, Transolver and MeshGraphNet on the BendVessel and ClotEntry datasets. Regarding single step performance, we found that Erwin performed well on BendVessel, while both Erwin and MeshGraphNet achieved good results for ClotEntry. On longer timeframes, we found that the loss of all models of stabilized for BendVessel, with MeshGraphNet stabilizing at lower loss values more often. In contrast, we found that the loss of all models quickly diverged for ClotEntry. Furthermore, we found that all three models provided significant speedups over traditional numerical solvers, with MeshGraphNet being the fastest. 

To improve single-step performance, we ran experiments using various data augmentations, namely rotations, datapoint size and data generation. We found that larger rotations improved performance, datapoint size has a marginal impact for our models and that a ratio of 1 : 2 real data to generated data was optimal. Lastly, we performed an ablation study that investigated the generalizability of Erwin on unseen rotations and found that generalization only occurred when the training data closely resembled the test data.


\section{Suitability as step-by-step surrogate models
for numerical mechanical thrombectomy simulations}
Our results of single-step performance on the test set showcased Erwins ability to perform well in both the small and medium data regimes. Erwin performed especially well on BendVessel compared to Transolver and MeshGraphNet, see table \ref{main_res}. One possible explanation of this fact is that Erwin can take almost full advantage of the attention mechanism at this lower datapoint size. This is because it performs full self-attention within each ball and the ball size (128) is of the same order of magnitude as the node count for BendVessel (768). When this is the case, the first BallProcessing layer of the encoder can encode each node with information of a large portion of the total nodes, as there are few balls each with relatively many nodes. This then results in the nodes in subsequent balls having more informative embeddings compared to when Erwin has to rely on aggregating information using its balltree, which is not a lossless process.

Furthermore, MeshGraphNet performed on par with Erwin for ClotEntry, but underperformed for BendVessel. This showcases that the best performing model can differ depending on the specifics of the dataset. Furthermore, it shows that MeshGraphNet could not take full advantage of this smaller datapoint size. This might be related to its use of the message-passing framework. This gives the model a certain receptive field size, which indicates how far apart (in terms of edges) the nodes can still interact. Since it performed well on the larger dataset we can assume it was large enough for that task. It then makes sense why the performance did not increase as much as Erwins when trained on the smaller BendVessel dataset. If the receptive field size was already large enough for the large task we would expect there to be no gain from decreasing the amount of nodes, as far apart nodes relevant to each other could already interact. 

Lastly, Transolver underperformed considerably for both datasets providing evidence that it is less suited to simulation emulation than the other models. Since both Erwin and Transolver are based on the transformer and Erwin did perform well this shows that the other parts of their architecture had a large impact. Since their main difference is how they apply attention, it could be the case that the approach of Transolver to divide the inputs into slices does not work well for 3D pointclouds. These slices are made globally (although they can use the spatial embeddings), while the balls that Erwin uses are explicitly made to be based on local position, which might explain this. However, Erwin and Transolver have many smaller design details, which each could explain this difference. 

In contrast to the single-step performance, the rollout results show less positive outcomes. While all models showed stable results for BendVessel, none of them performed well on ClotEntry across a longer time domain. Since these datasets differ mainly in the complexity of their geometry and in their node count, this indicates that one of these factors is important to stability. One possibility is that the vessel walls of ClotEntry are less smooth and regular than those of BendVessel, which makes rollout less forgiving as mistakes are quickly amplified. Interestingly, for BendVessel the loss of all models seemed to stabilize into two regions. Since the datapoints in BendVessel are very similar to each other, this could indicate that the models predict two separate outcomes, with one better matching the outcome of the numerical solver. If this is the case, then when using our methods in practice it would be useful to run the model multiple times and to determine if different outcomes occur as this would give insight into the models uncertainty.

Our results regarding the speed of our models show that all three models are capable of large speedups over traditional solvers. MeshGraphNet was the fastest and Erwin the slowest, with a difference of a factor of 4. However, since the speedups over traditional solvers were in the range of 80 - 400 times this difference is small in absolute rollout time, which was in the range of 0.5 - 13.2 seconds. This meant that in the worst case, Erwin was 10 seconds slower than MeshGraphNet. We do see this difference in absolute rollout time increase between BendVessel and ClotEntry, which indicates that this difference could become significant as the datapoints increase in size. This is supported by the fact that in general the speed of MeshGraphNets message passing architecture scales linearly with the number of nodes with constant node connectivity while attention scales quadratically. Of course, Erwin and Transolver try to mitigate this quadratic scaling as much as possible and thus comparing the scaling of these models is not straightforward, especially when considering that datasets with more nodes likely also require different hyperparameters for good performance.

Collectively, these findings suggest that both Erwin and MeshGraphnet are suitable as step-by-step surrogate models in the small data regime while Transolver is not. Erwin is very well suited to the small data regime, but as the size of the datapoints grow the speed advantage of MeshGraphNet becomes more significant. 
In the medium data regime, Erwin and MeshGraphNet show promising results as they are both fast and accurate for single steps. However, they currently miss the stability required to perform longer rollouts, making them not practically useful for this task as of yet.
Broadly, this means that as datasets increase in complexity we expect that more care needs to be taken to ensure stability. However, as we have seen, in terms of single-step performance and speed the methods, specifically Erwin and MeshGraphNet, presented in this thesis are likely well-suited to more complex domains.

\section{Impact of data augmentations on performance}
Firstly, the results of our experiments using rotations show that using no rotations, the differences in performance between Erwin, Transolver and MeshGraphNet are small. However, Erwin and MeshGraphNet benefit greatly from adding rotations to the training data. In contrast, Transolver seems to have its loss only decreased slightly when rotations are introduced. This suggests that one of the reasons that Transolver underperformed on the test set could be that it can not take advantage of the addition of rotational data augmentation. This is interesting as we used the same positional embeddings for Transolver and Erwin, which imply that the BallTree structure, and its breaking of rotational symmetry, of Erwin is one of its key contributing factors to its success using rotation augmentation.

Secondly, the evidence from our experiments varying the number of nodes indicates that all three models are robust to changes in node densities. After an initial increase in loss when the node count was too low, the loss stayed consistent when the amount of sampled nodes increased. This robustness to the node count shows that the hyperparameters, including the node connection radius, that were chosen generalize well to different node densities, as they were the same for all experiments.

Lastly, the results from our experiments using varying amounts of generated training data indicate that more generated data improves performance up to a point after which it has no impact. Furthermore, these improvements for all models show that BaryCentricSampling creates useful additional training datapoints. This implies that the new nodes that it creates on triangles represent the original nodes in a consistent way while they are different enough to still provide new information.

\section{Generalizability to unseen geometries}
The results from our experiment on generalizability show that Erwin did not generalize to an unseen rotation of the data, unless the training data contained datapoints had a rotation very similar to it. This suggests that in order for Erwin to perform well its training data must be adjusted to closely resemble the task it will be used on. In our experiments using rotations as data augmentations we saw that Erwin already performed well using $1.5$ radians of rotations on ClotEntry, which is not enough to transform every training datapoint into every direction (but is enough to rotate some to each direction as each datapoint is a vessel with different geometry). In combination, these findings suggest that while Erwin requires examples that closely match the test data, it can already perform well if only a portion of the training data does so.

Regarding our other models, since Transolver did not benefit from rotations nearly as much as Erwin did for our data augmentation experiment, it is possible that it would perform better on this generalizability task. This would for example be the case if the results from the data augmentation experiment were due to Transolver being partially rotation equivariant.
\section{Limitations}
The main limitation of this thesis is the completeness of the simulations that were used. They included large amounts of nodes, realistic vessel geometry and accurate physics simulation, but they did not include any devices explicitly nor did they use realistic blood clots (with different shapes and compositions). Simulations with explicit devices could prove challenging to emulate using the methods described in this thesis, especially if they use stent retrievers as these are complex devices with many small moving parts. However, since the simulations included realistic physics and many nodes, we argue that it still provides an adequate test to challenge our models.
Another limitation is that no extensive efficiency optimization was done. As such, the speed of the methods presented here might not scale well as the size of the datapoints increases, possibly rendering them impracticable. Furthermore, the methods used in this thesis mostly focused on single-step performance and not on rollout performance. As such, methods that specifically focus on longer timeframes were excluded from our analysis. Lastly, our methods did not take into account all the information provided by the physical nature of our simulation. Since the simulation was based on physical laws, the constraints of these laws (conservation of momentum etc.) are information that could be incorporated into our models, possibly improving performance and stability.

\section{Outlook}

 The methods presented in this thesis show promising results for the emulation of numerical physics simulations of mechanical thrombectomy using machine learning.
 However, how these methods will perform for more realistic simulations, those that include detailed devices and more nodes, is still not clear. Thus, before it can be considered how machine learning models can be used practically for this problem, future research must investigate how these methods perform on realistic simulations. 

Future research could perform experiments on some of the simulations that are already avaliable today, as they already include detailed thrombectomy devices. This is currently challenging as most simulations that are presented in research papers are not made available as datasets with machine learning in mind. Instead the data needed for training machine learning models is discarded as soon as analysis is done. 

Furthermore, for our methods, stability over longer time periods is a problem for more complicated datasets. This poses another challenge for the practical application of these methods as generating rollouts is their most interesting use case. Future research could incorporate existing methods for improving stability such as training on multi-step rollouts or use machine learning architectures that are specifically made for increased rollout stability, such as LE-PDE \cite{wu2022learning}.

To improve performance even more, further research could focus on finding priors that can be added to the loss during training. It is possible that useful priors exist that can take advantage of the specific dynamics present in mechanical thrombectomy. An example of such a prior is that the distances between adjacent blood clot nodes largely stay the same across timesteps, as blood clots are not fluid and thus nodes can not move freely throughout the clot. Additionally, since this is a known physical system, physics based constraints can be used, such as those used in Physics-Informed Neural Networks \cite{raissi2019physics}.

As the quality of simulations improve their usefulness in clinical settings will increase. As such, the potential for the use of machine learning methods to speed them up increases as well. The methods we have presented are a first step to this use-case, as they show how machine learning methods can be used to speed up simplified numerical simulations of thrombectomy step-by-step. If these methods are eventually improved enough they might enable the practical use of increasingly more complicated simulations in time-sensitive clinical situations. In the long term, methods such as ours can contribute to the use of fast emulations of patient-specific numerical simulations for ischemic stroke patients, giving operators more insight into the potential results of their procedures. While challenges such as rollout stability and scaling to large datasets remain, the findings of this thesis indicate that machine learning based step-by-step surrogate models provide a promising approach for speeding up numerical physics simulations of mechanical thrombectomy and thus of their use in clinical settings.

\section{Conclusion}

This thesis investigated the suitability of step-by-step machine learning surrogate models for accelerating numerical simulations of mechanical thrombectomy. 
We trained three machine learning models on two datasets that involved simplified mechanical thrombectomy procedures, showing that two models achieved high single-step accuracy and that all models provided a significant speedup over traditional solvers. However, for multi-step rollout, stable results were only observed for simple geometries, while rollouts on complex geometries were unstable for all models, showcasing the need for techniques that provide multi-step stability.
 Additionally, we investigated the impact of data augmentations and found that they were critical to model performance. Namely, we found that larger rotations and the addition of generated data improved performance, while datapoint size had a marginal impact for our models. Lastly, we investigated the generalizability of one of our models to unseen geometries and found that generalization requires a close match between training and testing data, underscoring the important role of data augmentation in the performance of surrogate models.
Overall, our findings highlight the potential for machine learning based step-by-step surrogate models to speed up numerical thrombectomy simulations and provides a foundation for future studies to develop stable methods that scale to realistic simulations. 
\appendix
\chapter{Hyperparameters}\label{hyper}
Any hyperparameters not mentioned here were left at their default values.
\section{Erwin}

\begin{table}[ht]
\centering
\begin{tabular}{|l|l|}
\hline
\textbf{Parameter} & \textbf{Value} \\
\hline
Batch size & 8 \\
Radius & 0.766 \\
Hidden dimension & 32 \\
Encoder heads & {[}4, 4, 4, 4{]} \\
Encoder depths & {[}8, 10, 12, 14{]} \\
Decoder heads & {[}4, 4, 4{]} \\
Decoder depths & {[}4, 4, 4{]} \\
Strides & {[}2, 2, 2{]} \\
Ball sizes & {[}128, 128, 128, 128{]} \\
MLP ratio & 4 \\
Rotate & 45 \\
Optimizer & AdamW \\
LR scheduler & Linear Warmup + Cosine Decay \\
Learning rate & 1.00E--04 \\
Warmup epochs & 10 \\
Total epochs & 300 \\
\hline
\end{tabular}
\caption{Hyperparameter settings for Erwin}
\label{tab:hyperparams_clot_entry_vertical}
\end{table}

\section{Transolver}
\begin{table}[H]
\centering
\small
\begin{tabular}{|l|l|}
\hline
\textbf{Parameter} & \textbf{Value} \\
\hline
Batch size & 16 \\
Radius & 0.576 \\
Hidden dimension & 64 \\
Learning rate & 1.00E--04 \\
Dropout & 0.576 \\
Number of heads & 16 \\
Number of layers & 5 \\
Slice number & 24 \\
Optimizer & AdamW \\
LR scheduler & Linear Warmup + Cosine Decay \\
Warmup epochs & 10 \\
Total epochs & 300 \\
\hline
\end{tabular}
\caption{Hyperparameter settings for Transolver}
\label{tab:hyperparams_clot_entry_alt}
\end{table}
\section{MeshGraphNet}
\begin{table}[h]
\centering
\small
\begin{tabular}{|l|l|}
\hline
\textbf{Parameter} & \textbf{Value} \\
\hline
Batch size & 8 \\
Radius & 0.56 \\
Hidden dimension & 64 \\
Learning rate & 10E--04 \\
Processor size & 15 \\
Optimizer & AdamW \\
LR scheduler & Linear Warmup + Cosine Decay \\
Warmup epochs & 10 \\
Total epochs & 300 \\
\hline
\end{tabular}
\caption{Hyperparameter settings for MeshGraphNet}
\label{tab:hyperparams_clot_entry_third}
\end{table}

\section{Learning Rate Scheduler}
\begin{figure}[h]
    \centering
    \includegraphics[width=0.5\linewidth]{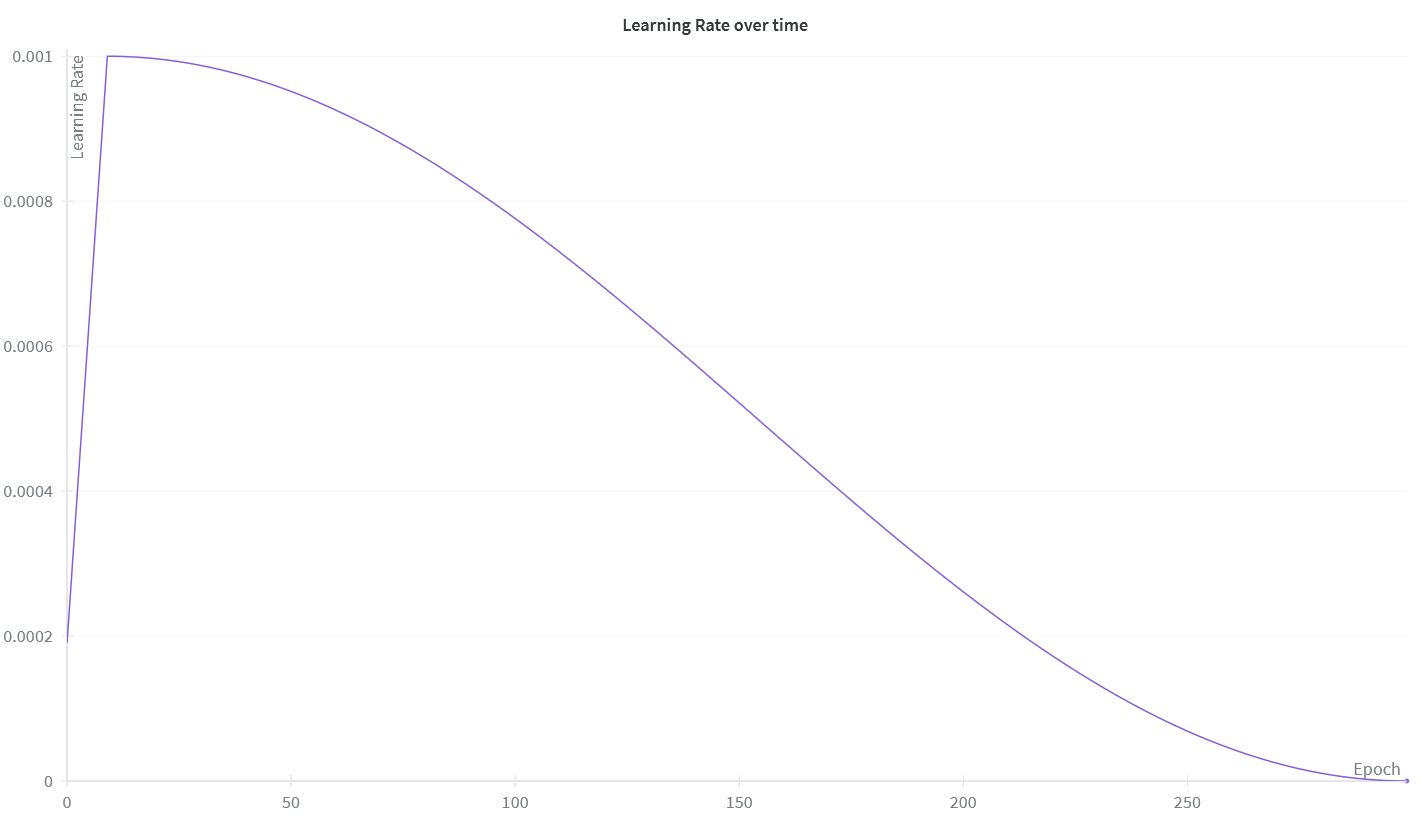}
    \caption{Example learning rate schedule, with 300 epochs, a learning rate value of 0.001 and using a linear warmup of 10 epochs + Cosine Decay}
    \label{fig:learningrate}
\end{figure}

\bibliographystyle{unsrt}
\bibliography{bibentries}

@article{fan2023global,
  title={Global burden, risk factor analysis, and prediction study of ischemic stroke, 1990--2030},
  author={Fan, Jiahui and Li, Xiaoguang and Yu, Xueying and Liu, Zhenqiu and Jiang, Yanfeng and Fang, Yibin and Zong, Ming and Suo, Chen and Man, Qiuhong and Xiong, Lize},
  journal={Neurology},
  volume={101},
  number={2},
  pages={e137--e150},
  year={2023},
  publisher={Lippincott Williams \& Wilkins Hagerstown, MD}
}

@article{jansen2018endovascular,
  title={Endovascular treatment for acute ischaemic stroke in routine clinical practice: prospective, observational cohort study (MR CLEAN Registry)},
  author={Jansen, Ivo GH and Mulder, Maxim JHL and Goldhoorn, Robert-Jan B},
  journal={bmj},
  volume={360},
  year={2018},
  publisher={British Medical Journal Publishing Group}
}

@article{wang2310discovering,
  title={Discovering symmetry breaking in physical systems with relaxed group convolution, 2024},
  author={Wang, Rui and Hofgard, Elyssa and Gao, Han and Walters, Robin and Smidt, Tess E},
  journal={URL https://arxiv. org/abs/2310.02299},
  year={2023}
}

@article{walters2020applications,
  title={Applications of deep learning in molecule generation and molecular property prediction},
  author={Walters, W Patrick and Barzilay, Regina},
  journal={Accounts of chemical research},
  volume={54},
  number={2},
  pages={263--270},
  year={2020},
  publisher={ACS Publications}
}

@article{mosconi2022treatments,
  title={Treatments in ischemic stroke: current and future},
  author={Mosconi, Maria Giulia and Paciaroni, Maurizio},
  journal={European Neurology},
  volume={85},
  number={5},
  pages={349--366},
  year={2022},
  publisher={S. Karger AG}
}

@article{rizzo1967integral,
  title={An integral equation approach to boundary value problems of classical elastostatics},
  author={Rizzo, Frank J},
  journal={Quarterly of applied mathematics},
  volume={25},
  number={1},
  pages={83--95},
  year={1967}
}

@article{belytschko1984explicit,
  title={Explicit algorithms for the nonlinear dynamics of shells},
  author={Belytschko, Ted and Lin, Jerry I and Chen-Shyh, Tsay},
  journal={Computer methods in applied mechanics and engineering},
  volume={42},
  number={2},
  pages={225--251},
  year={1984},
  publisher={Elsevier}
}

@article{chen2011multiphysics,
  title={Multiphysics simulations of rocket engine combustion},
  author={Chen, Yen-Sen and Chou, TH and Gu, BR and Wu, Jong-Shinn and Wu, Bill and Lian, YY and Yang, Luke},
  journal={Computers \& Fluids},
  volume={45},
  number={1},
  pages={29--36},
  year={2011},
  publisher={Elsevier}
}

@book{langanke1991computational,
  title={Computational nuclear physics},
  author={Langanke, Karlheinz and Maruhn, Joachim A and Koonin, Steven E},
  year={1991},
  publisher={Springer}
}

@article{fumagalli2024role,
  title={The role of computational methods in cardiovascular medicine: a narrative review},
  author={Fumagalli, Ivan and Pagani, Stefano and Vergara, Christian and Adebo, Dilachew A and Del Greco, Maurizio and Frontera, Antonio and Luciani, Giovanni Battista and Pontone, Gianluca and Scrofani, Roberto and Quarteroni, Alfio and others},
  journal={Translational Pediatrics},
  volume={13},
  number={1},
  pages={146},
  year={2024}
}

@article{anand2021trends,
  title={Trends in acute ischemic stroke treatments and mortality in the United States from 2012 to 2018},
  author={Anand, Sharath Kumar and Benjamin, William J and Adapa, Arjun Rohit and Park, Jiwon V and Wilkinson, D Andrew and Daou, Badih J and Burke, James F and Pandey, Aditya S},
  journal={Neurosurgical focus},
  volume={51},
  number={1},
  pages={E2},
  year={2021},
  publisher={American Association of Neurological Surgeons}
}

@phdthesis{daei2023computational,
  title={Computational modeling of interventional and implantable endovascular devices},
  author={Daei, Mahdi Daei},
  year={2023},
  school={Institut Polytechnique de Paris}
}

@inproceedings{gori2005new,
  title={A new model for learning in graph domains},
  author={Gori, Marco and Monfardini, Gabriele and Scarselli, Franco},
  booktitle={Proceedings. 2005 IEEE international joint conference on neural networks, 2005.},
  volume={2},
  pages={729--734},
  year={2005},
  organization={IEEE}
}

@inproceedings{pfaff2020learning,
  title={Learning mesh-based simulation with graph networks},
  author={Pfaff, Tobias and Fortunato, Meire and Sanchez-Gonzalez, Alvaro and Battaglia, Peter},
  booktitle={International conference on learning representations},
  year={2020}
}

@article{TIMI1985,
  author    = {{TIMI Study Group}},
  title     = {The Thrombolysis in Myocardial Infarction (TIMI) trial. Phase I findings},
  journal   = {The New England Journal of Medicine},
  year      = {1985},
  month     = {April 4},
  volume    = {312},
  number    = {14},
  pages     = {932--936},
  doi       = {10.1056/NEJM198504043121437},
  pmid      = {4038784}
}

@article{liang2018deep,
  title={A deep learning approach to estimate stress distribution: a fast and accurate surrogate of finite-element analysis},
  author={Liang, Liang and Liu, Minliang and Martin, Caitlin and Sun, Wei},
  journal={Journal of The Royal Society Interface},
  volume={15},
  number={138},
  pages={20170844},
  year={2018},
  publisher={The Royal Society}
}

@article{liang2017machine,
  title={A machine learning approach to investigate the relationship between shape features and numerically predicted risk of ascending aortic aneurysm},
  author={Liang, Liang and Liu, Minliang and Martin, Caitlin and Elefteriades, John A and Sun, Wei},
  journal={Biomechanics and modeling in mechanobiology},
  volume={16},
  number={5},
  pages={1519--1533},
  year={2017},
  publisher={Springer}
}

@inproceedings{kuipers2024generating,
  title={Generating cerebral vessel trees of acute ischemic stroke patients using conditional set-diffusion},
  author={Kuipers, Thijs P and Konduri, Praneeta R and Marquering, Henk and Bekkers, Erik J},
  booktitle={Medical Imaging with Deep Learning},
  year={2024}
}

@article{wu2402transolver,
  title={Transolver: A Fast Transformer Solver for PDEs on General Geometries (2024)},
  author={Wu, Haixu and Luo, Huakun and Wang, Haowen and Wang, Jianmin and Long, Mingsheng},
  journal={URL https://arxiv. org/abs/2402.02366},
  year={2024}
}

@article{munich2019overview,
  title={Overview of mechanical thrombectomy techniques},
  author={Munich, Stephan A and Vakharia, Kunal and Levy, Elad I},
  journal={Neurosurgery},
  volume={85},
  number={suppl\_1},
  pages={S60--S67},
  year={2019},
  publisher={LWW}
}

@misc{penumbraACE,
  author       = {Penumbra Inc.},
  title        = {Penumbra ACE},
  howpublished = {\url{https://www.penumbrainc.com/products/penumbra-ace/}},
  note         = {Accessed: 2025-11-28}
}

@misc{medtronicReactCatheter,
  author       = {Medtronic},
  title        = {React Aspiration Catheter},
  howpublished = {\url{https://www.medtronic.com/en-us/healthcare-professionals/products/neurological/neurovascular/catheters/aspiration-neurovascular-catheters/react-catheter.html}},
  note         = {Accessed: 2025-11-28}
}

@misc{strykerTrevoXP,
  author       = {Stryker},
  title        = {Trevo XP ProVue Retriever},
  howpublished = {\url{https://www.stryker.com/us/en/neurovascular/products/trevo--xp-provue-retriever-.html}},
  note         = {Accessed: 2025-11-28}
}

@misc{terumoERIC,
  author       = {Terumo Neurovascular},
  title        = {ERIC Device},
  howpublished = {\url{https://www.terumoneuro.com/emea/products/eric}},
  note         = {Accessed: 2025-11-28}
}

@article{zaidat2018first,
  title={First pass effect: a new measure for stroke thrombectomy devices},
  author={Zaidat, Osama O and Castonguay, Alicia C and Linfante, Italo and Gupta, Rishi and Martin, Coleman O and Holloway, William E and Mueller-Kronast, Nils and English, Joey D and Dabus, Guilherme and Malisch, Tim W and others},
  journal={Stroke},
  volume={49},
  number={3},
  pages={660--666},
  year={2018},
  publisher={Lippincott Williams \& Wilkins Hagerstown, MD}
}

@article{kaesmacher2017risk,
  title={Risk of thrombus fragmentation during endovascular stroke treatment},
  author={Kaesmacher, J and Boeckh-Behrens, T and Simon, S and Maegerlein, C and Kleine, JF and Zimmer, C and Schirmer, L and Poppert, H and Huber, T},
  journal={American Journal of Neuroradiology},
  volume={38},
  number={5},
  pages={991--998},
  year={2017},
  publisher={American Journal of Neuroradiology}
}

@article{boodt2021mechanical,
  title={Mechanical characterization of thrombi retrieved with endovascular thrombectomy in patients with acute ischemic stroke},
  author={Boodt, Nikki and Snouckaert van Schauburg, Philip RW and Hund, Hajo M and Fereidoonnezhad, Behrooz and McGarry, J Patrick and Akyildiz, Ali C and Van Es, Adriaan CGM and De Meyer, Simon F and Dippel, Diederik WJ and Lingsma, Hester F and others},
  journal={Stroke},
  volume={52},
  number={8},
  pages={2510--2517},
  year={2021},
  publisher={Lippincott Williams \& Wilkins Hagerstown, MD}
}

@article{yeo2019does,
  title={Why does mechanical thrombectomy in large vessel occlusion sometimes fail? A review of the literature},
  author={Yeo, Leonard LL and Bhogal, Pervinder and Gopinathan, Anil and Cunli, Yang and Tan, Benjamin and Andersson, Tommy},
  journal={Clinical Neuroradiology},
  volume={29},
  number={3},
  pages={401--414},
  year={2019},
  publisher={Springer}
}

@article{kapoor2008variations,
  title={Variations in the configuration of the circle of Willis},
  author={Kapoor, Kanchan and Singh, Balbir and Dewan, Inder Jit},
  journal={Anatomical science international},
  volume={83},
  number={2},
  pages={96--106},
  year={2008},
  publisher={Springer}
}

@article{luraghi2021first,
  title={The first virtual patient-specific thrombectomy procedure},
  author={Luraghi, Giulia and Bridio, Sara and Matas, Jose Felix Rodriguez and Dubini, Gabriele and Boodt, Nikki and Gijsen, Frank JH and van der Lugt, Aad and Fereidoonnezhad, Behrooz and Moerman, Kevin M and McGarry, Patrick and others},
  journal={Journal of Biomechanics},
  volume={126},
  pages={110622},
  year={2021},
  publisher={Elsevier}
}

@article{shazeer2020glu,
  title={Glu variants improve transformer},
  author={Shazeer, Noam},
  journal={arXiv preprint arXiv:2002.05202},
  year={2020}
}

@article{musuka2015diagnosis,
  title={Diagnosis and management of acute ischemic stroke: speed is critical},
  author={Musuka, Tapuwa D and Wilton, Stephen B and Traboulsi, Mouhieddin and Hill, Michael D},
  journal={Cmaj},
  volume={187},
  number={12},
  pages={887--893},
  year={2015},
  publisher={CMAJ}
}

@article{liu2022simulation,
  title={Simulation of stent retriever thrombectomy in acute ischemic stroke by finite element analysis},
  author={Liu, Ronghui and Jin, Chang and Wang, Lizhen and Yang, Yisong and Fan, Yubo and Wang, Weidong},
  journal={Computer Methods in Biomechanics and Biomedical Engineering},
  volume={25},
  number={7},
  pages={740--749},
  year={2022},
  publisher={Taylor \& Francis}
}

@inproceedings{ronneberger2015u,
  title={U-net: Convolutional networks for biomedical image segmentation},
  author={Ronneberger, Olaf and Fischer, Philipp and Brox, Thomas},
  booktitle={International Conference on Medical image computing and computer-assisted intervention},
  pages={234--241},
  year={2015},
  organization={Springer}
}

@article{wortmann2022development,
  title={Development of synthetic thrombus models to simulate stroke treatment in a physical neurointerventional training model},
  author={Wortmann, Nadine and Andersek, Thomas and Guerreiro, Helena and Kyselyova, Anna A and Fr{\"o}lich, Andreas M and Fiehler, Jens and Krause, Dieter},
  journal={All Life},
  volume={15},
  number={1},
  pages={283--301},
  year={2022},
  publisher={Taylor \& Francis}
}

@article{groves2023review,
  title={A review of low-cost ultrasound compatible phantoms},
  author={Groves, Leah A and Keita, Mohamed and Talla, Saidou and Kikinis, Ron and Fichtinger, Gabor and Mousavi, Parvin and Camara, Mamadou},
  journal={IEEE Transactions on Biomedical Engineering},
  volume={70},
  number={12},
  pages={3436--3448},
  year={2023},
  publisher={IEEE}
}

@article{hemmasian2024multi,
  title={Multi-scale time-stepping of Partial Differential Equations with transformers},
  author={Hemmasian, AmirPouya and Farimani, Amir Barati},
  journal={Computer Methods in Applied Mechanics and Engineering},
  volume={426},
  pages={116983},
  year={2024},
  publisher={Elsevier}
}

@article{groan2021time,
  title={Time-based decision making for reperfusion in acute ischemic stroke},
  author={Gr{\o}an, Mathias and Ospel, Johanna and Ajmi, Soffien and Sandset, Else Charlotte and Kurz, Martin W and Skjelland, Mona and Advani, Rajiv},
  journal={Frontiers in Neurology},
  volume={12},
  pages={728012},
  year={2021},
  publisher={Frontiers Media SA}
}

@inproceedings{li2021neural,
  title={Neural operator: Learning maps between function spaces},
  author={Li, Zongyi},
  booktitle={2021 Fall Western Sectional Meeting},
  year={2021},
  organization={AMS}
}

@article{gg2018acute,
  title={Acute ischemic stroke: a review of imaging, patient selection, and management in the endovascular era. Part II: Patient selection, endovascular thrombectomy, and postprocedure management},
  author={GG, Sharath Kumar and Nagesh, Chinmay P},
  journal={Journal of Clinical Interventional Radiology ISVIR},
  volume={2},
  number={03},
  pages={169--183},
  year={2018},
  publisher={Thieme Medical and Scientific Publishers Private Ltd.}
}

@article{zhang2019root,
  title={Root mean square layer normalization},
  author={Zhang, Biao and Sennrich, Rico},
  journal={Advances in neural information processing systems},
  volume={32},
  year={2019}
}

@inproceedings{dauphin2017language,
  title={Language modeling with gated convolutional networks},
  author={Dauphin, Yann N and Fan, Angela and Auli, Michael and Grangier, David},
  booktitle={International conference on machine learning},
  pages={933--941},
  year={2017},
  organization={PMLR}
}

@article{zhdanov2025erwin,
  title={Erwin: A tree-based hierarchical transformer for large-scale physical systems},
  author={Zhdanov, Maksim and Welling, Max and van de Meent, Jan-Willem},
  journal={arXiv preprint arXiv:2502.17019},
  year={2025}
}

@article{di2019acute,
  title={Acute ischemic stroke thrombi have an outer shell that impairs fibrinolysis},
  author={Di Meglio, Lucas and Desilles, Jean-Philippe and Ollivier, V{\'e}ronique and Nomenjanahary, Mialitiana Solo and Di Meglio, Sara and Deschildre, Catherine and Loyau, St{\'e}phane and Olivot, Jean-Marc and Blanc, Rapha{\"e}l and Piotin, Michel and others},
  journal={Neurology},
  volume={93},
  number={18},
  pages={e1686--e1698},
  year={2019},
  publisher={Lippincott Williams \& Wilkins Hagerstown, MD}
}

@article{jolugbo2021thrombus,
  title={Thrombus composition and efficacy of thrombolysis and thrombectomy in acute ischemic stroke},
  author={Jolugbo, Precious and Ariens, Robert AS},
  journal={Stroke},
  volume={52},
  number={3},
  pages={1131--1142},
  year={2021},
  publisher={Lippincott Williams \& Wilkins Hagerstown, MD}
}

@article{onal2020feasibility,
  title={Feasibility of distal mechanical thrombectomy in M3, A3 and P3 segments via a 0.013-inch delivery system: preliminary experience},
  author={Onal, Yilmaz and Velioglu, Murat and Demir, Ugur and Celikoglu, Erhan and Karakas, Hakki M},
  journal={Turk Neurosurg},
  volume={30},
  number={4},
  pages={614--620},
  year={2020}
}

@article{jadhav2021indications,
  title={Indications for mechanical thrombectomy for acute ischemic stroke: current guidelines and beyond},
  author={Jadhav, Ashutosh P and Desai, Shashvat M and Jovin, Tudor G},
  journal={Neurology},
  volume={97},
  number={20\_Supplement\_2},
  pages={S126--S136},
  year={2021},
  publisher={Lippincott Williams \& Wilkins Hagerstown, MD}
}

@article{wu2022learning,
  title={Learning to accelerate partial differential equations via latent global evolution},
  author={Wu, Tailin and Maruyama, Takashi and Leskovec, Jure},
  journal={Advances in Neural Information Processing Systems},
  volume={35},
  pages={2240--2253},
  year={2022}
}

@article{vaswani2017attention,
  title={Attention is all you need},
  author={Vaswani, Ashish and Shazeer, Noam and Parmar, Niki and Uszkoreit, Jakob and Jones, Llion and Gomez, Aidan N and Kaiser, {\L}ukasz and Polosukhin, Illia},
  journal={Advances in neural information processing systems},
  volume={30},
  year={2017}
}

@inproceedings{gilmer2017neural,
  title={Neural message passing for quantum chemistry},
  author={Gilmer, Justin and Schoenholz, Samuel S and Riley, Patrick F and Vinyals, Oriol and Dahl, George E},
  booktitle={International conference on machine learning},
  pages={1263--1272},
  year={2017},
  organization={Pmlr}
}

@misc{Alammar2026,
  author       = {Jay Alammar},
  title        = {The Illustrated Transformer},
  howpublished = {\url{https://jalammar.github.io/illustrated-transformer/}},
  year         = {2018},
  note         = {Accessed: 2026-01-31; originally published June 27, 2018} 
}

@article{schomer1994anatomy,
  title={The anatomy of the posterior communicating artery as a risk factor for ischemic cerebral infarction},
  author={Schomer, Don F and Marks, Michael P and Steinberg, Gary K and Johnstone, Iain M and Boothroyd, Derek B and Ross, Michael R and Pelc, Norbert J and Enzmann, Dieter R},
  journal={New England Journal of Medicine},
  volume={330},
  number={22},
  pages={1565--1570},
  year={1994},
  publisher={Mass Medical Soc}
}

@article{raissi2019physics,
  title={Physics-informed neural networks: A deep learning framework for solving forward and inverse problems involving nonlinear partial differential equations},
  author={Raissi, Maziar and Perdikaris, Paris and Karniadakis, George E},
  journal={Journal of Computational physics},
  volume={378},
  pages={686--707},
  year={2019},
  publisher={Elsevier}
}

@article{beumer2016occurrence,
  title={Occurrence of intracranial large vessel occlusion in consecutive, non-referred patients with acute ischemic stroke},
  author={Beumer, Debbie and Mulder, Maxim JHL and Saiedie, Ghesrouw and Fonville, Susanne and van Oostenbrugge, Robert J and van Zwam, Wim H and Homburg, Philip J and van der Lugt, Aad and Dippel, Diederik WJ},
  journal={Neurovascular Imaging},
  volume={2},
  number={1},
  pages={11},
  year={2016},
  publisher={Springer}
}

@article{sturiale2024geometry,
  title={Geometry and Symmetry of Willis’ Circle and Middle Cerebral Artery Aneurysms Development},
  author={Sturiale, Carmelo Lucio and Scerrati, Alba and Ricciardi, Luca and Rustemi, Oriela and Auricchio, Anna Maria and Norri, Nicol{\`o} and Piazza, Amedeo and Raneri, Fabio and Benato, Alberto and Albanese, Alessio and others},
  journal={Journal of Clinical Medicine},
  volume={13},
  number={10},
  pages={2808},
  year={2024},
  publisher={MDPI}
}

@article{regazzoni2024learning,
  title={Learning the intrinsic dynamics of spatio-temporal processes through Latent Dynamics Networks},
  author={Regazzoni, Francesco and Pagani, Stefano and Salvador, Matteo and Dede’, Luca and Quarteroni, Alfio},
  journal={Nature Communications},
  volume={15},
  number={1},
  pages={1834},
  year={2024},
  publisher={Nature Publishing Group UK London}
}

@article{nishi2019predicting,
  title={Predicting clinical outcomes of large vessel occlusion before mechanical thrombectomy using machine learning},
  author={Nishi, Hidehisa and Oishi, Naoya and Ishii, Akira and Ono, Isao and Ogura, Takenori and Sunohara, Tadashi and Chihara, Hideo and Fukumitsu, Ryu and Okawa, Masakazu and Yamana, Norikazu and others},
  journal={Stroke},
  volume={50},
  number={9},
  pages={2379--2388},
  year={2019},
  publisher={Lippincott Williams \& Wilkins Hagerstown, MD}
}

@article{loshchilov2017decoupled,
  title={Decoupled weight decay regularization},
  author={Loshchilov, Ilya and Hutter, Frank},
  journal={arXiv preprint arXiv:1711.05101},
  year={2017}
}

@misc{physicsnemo2023,
  title        = {NVIDIA PhysicsNeMo: An Open-Source Framework for Physics-Based Deep Learning in Science and Engineering},
  author       = {{PhysicsNeMo Contributors}},
  year         = {2023},
  month        = feb,
  day          = {24},
  url          = {https://github.com/NVIDIA/physicsnemo},
  note         = {Accessed: 2026-01-21}
}

@article{blanc2020recent,
  title={Recent advances in devices for mechanical thrombectomy},
  author={Blanc, Rapha{\"e}l and Escalard, Simon and Baharvadhat, Humain and Desilles, Jean Philippe and Boisseau, William and Fahed, Robert and Redjem, Hocine and Ciccio, Gabriele and Smajda, Stanislas and Maier, Benjamin and others},
  journal={Expert review of medical devices},
  volume={17},
  number={7},
  pages={697--706},
  year={2020},
  publisher={Taylor \& Francis}
}

@article{Lapergue2017ASTER,
  author    = {Lapergue, B. and Blanc, R. and Gory, B. and others},
  title     = {Effect of Endovascular Contact Aspiration vs Stent Retriever on Revascularization in Patients With Acute Ischemic Stroke and Large Vessel Occlusion: The ASTER Randomized Clinical Trial},
  journal   = {JAMA},
  year      = {2017},
  volume    = {318},
  number    = {5},
  pages     = {443--452},
  doi       = {10.1001/jama.2017.9644}
}
\end{document}